\documentclass[journal]{IEEEtran}
\usepackage{graphicx}
\usepackage{caption,subcaption}
\usepackage{amsmath}
\usepackage{amssymb}
\usepackage{booktabs}
\usepackage[ruled]{algorithm2e}

\usepackage[capitalize]{cleveref}
\crefname{section}{Sec.}{Secs.}
\Crefname{section}{Section}{Sections}
\Crefname{table}{Table}{Tables}
\crefname{table}{Tab.}{Tabs.}

\begin{document}
%

\title{GaitFi: Robust Device-Free Human Identification via WiFi and Vision Multimodal Learning}

%
%
%

\author{Lang~Deng$^*$,
    Jianfei~Yang$^*$,
    Shenghai~Yuan,
	Han~Zou,
	Chris~Xiaoxuan~Lu,
 	and~Lihua~Xie,~\IEEEmembership{Fellow,~IEEE}
 
 \thanks{
	L. Deng, J. Yang, S. Yuan and L. Xie are with the School of Electrical and Electronics Engineering, Nanyang Technological University, Singapore (e-mail: ldeng002@e.ntu.edu.sg; yang0478@e.ntu.edu.sg; syuan003@e.ntu.edu.sg; elhxie@ntu.edu.sg).
 
 	H. Zou is with the Department of Electrical Engineering and Computer Sciences, University of California, Berkeley, USA (e-mail: enthalpyzou@gmail.com).
 	
 	C. X. Lu is with the School of Informatics at the University of Edinburgh, United Kingdom (e-mail: xiaoxuan.lu@ed.ac.uk).
 	
 	$^*$These authors contributed equally to this work. J. Yang is the corresponding author (e-mail: yang0478@e.ntu.edu.sg).
	
	This work is supported by NTU Presidential Postdoctoral Fellowship, ``Adaptive Multimodal Learning for Robust Sensing and Recognition in Smart Cities'' project fund, in Nanyang Technological University, Singapore.
	}
}

%
%

\markboth{}%
{}
%



\maketitle

\begin{abstract}
As an important biomarker for human identification, human gait can be collected at a distance by passive sensors without subject cooperation, which plays an essential role in crime prevention, security detection and other human identification applications. At present, most research works are based on cameras and computer vision techniques to perform gait recognition. However, vision-based methods are not reliable when confronting poor illuminations, leading to degrading performances. In this paper, we propose a novel multimodal gait recognition method, namely GaitFi, which leverages WiFi signals and videos for human identification. In GaitFi, Channel State Information (CSI) that reflects the multi-path propagation of WiFi is collected to capture human gaits, while videos are captured by cameras. To learn robust gait information, we propose a Lightweight Residual Convolution Network (LRCN) as the backbone network, and further propose the two-stream GaitFi by integrating WiFi and vision features for the gait retrieval task. The GaitFi is trained by the triplet loss and classification loss on different levels of features. Extensive experiments are conducted in the real world, which demonstrates that the GaitFi outperforms state-of-the-art gait recognition methods based on single WiFi or camera, achieving 94.2\% for human identification tasks of 12 subjects. 

\end{abstract}


%
\IEEEpeerreviewmaketitle

\section{Introduction}
\label{sec:intro}

\IEEEPARstart{N}{owadays}, numerous intelligent monitoring systems have been deployed in the public domain to extract biomarker information related to human behavior and identities. With the development of Internet of Things (IoT) sensors and pattern recognition, various human identification technologies come into existence, such as fingerprint recognition~\cite{ali2016overview}, iris recognition~\cite{de2016iris} and face recognition~\cite{8614364}. Though these technologies achieve remarkable performances, they still have their own limitations, such as the sensing range of fingerprint and iris recognition and the degrading face recognition due to the mask during the COVID-19 period~\cite{wang2021mask}. Different from the prevailing human identification methods, gait is a unique biomarker that can be identified at a distance without human cooperation. The advantages of gait recognition in remote monitoring~\cite{bouchrika2018survey} make it essential in crime prevention, forensic identification and social security.

Human gait is defined as the coordinated and cyclic combination of various movements in the walking action that is a unique biomarker for a person~\cite{isaac2019trait}. Specifically, the gait information includes the static information of the individual's appearance and the dynamic information of the person's walking. Therefore, gait recognition can be achieved by extracting these interrelated salient features. The advantage of gait-based human identification is that gait can not only be captured at a longer distance but is also difficult to imitate, which has attracted many researchers to employ various sensors for gait recognition~\cite{yang2019review,zou2018identification}.

Existing gait recognition methods mainly rely on cameras~\cite{kumar2021gait}, wearable devices~\cite{yang2019review}, and radar~\cite{vandersmissen2018indoor}. However, they have distinct limitations due to the characteristics of each modality of sensors. For cameras-based solutions, the video can be easily affected by environmental conditions such as illumination and occlusion. Some scenarios even forbid the camera to be used due to user privacy. Wearable devices can be leveraged for human activity~\cite{ding2018energy} and gait analytics~\cite{marsico2019survey}, which allows for higher resolution measurements using multiple sensors, but wearable devices require the cooperation of subjects, which restrains their application scenarios, such as crime prevention. Many radar-based methods have also been utilized for gait recognition~\cite{chen2021attention}, which can extract gait information by utilizing the Doppler feature of Frequency-Modulated Continuous Wave (FMCW). However, the disadvantage of radar is the sparsity of the data with low SNR. Conversely, Lidar can obtain higher-resolution data than radar, but it is very expensive~\cite{benedek2016lidar}. Recently, WiFi is enabled to sense human gaits by extracting channel state information (CSI)~\cite{wang2016gait,yang2022efficientfi}, which is proved to be cost-effective and privacy-preserving. In WiFi sensing, human gaits are reflected by detailed amplitude and phase information of different subcarriers after the WiFi signals are modulated by Orthogonal Frequency Division Multiplexing (OFDM)~\cite{xie2015precise}. Since the body motions interfere with the propagation path of the Wifi signals and the body motions of the gaits of different subjects are different, these lead to specific patterns in CSI for different subjects. Since each sensor modality has its pros and cons, is it possible to fuse a few complementary modalities for robust gait recognition? 

Here we consider the most common modality, video camera that contains a large amount of information, and the WiFi. The reason for choosing WiFi as another modality in addition to visual cameras is that WiFi-enabled IoT devices are more ubiquitous when compared to lidar and radar. The CSI data extracted from WiFi is robust to illumination, which is a good complementarity to vision modality. As WiFi sensing leverages electromagnetic waves rather than visible light, when slight occlusion happens for the camera (e.g., plastic and paper materials), the WiFi system can still work, which denotes another merit.

In this paper, we propose a multimodal device-free human identification system utilizing the gait recognition method, namely GaitFi, which can recognize human identities based on commercial off-the-shelf (COTS) WiFi-enabled IoT devices and cameras. GaitFi consists of a two-stream network for WiFi and video, and a multimodal fusion module to recognize human gaits. For the WiFi sensing module, we propose a Lightweight Residual Convolution Network (LRCN) that consists of convolution layers and residual blocks to extract spatial and temporal features. For the vision sensing module, we first use LRCN to obtain frame-level features and a Long Short-Term Memory (LSTM) network~\cite{yu2019review} to extract temporal dynamics. In the modality fusion module, we concatenate the feature vectors from the two modalities to generate a robust gait representation. This concatenated feature vector is mapped to the prediction probability. We apply the cross-entropy loss and the triplet loss to GaitFi on the final prediction and the concatenated feature layers, respectively, so that the two losses contribute to robust classification and metric learning feature space without interference. By using lightweight backbone and uncomplicated multmodal learning framework, GaitFi achieves good human identification performance with relatively small computational complexity and relatively small inference time. To demonstrate the effectiveness, we conduct real-world experiments by implementing the system using a pair of WiFi routers, a camera and a mini-PC. The proposed GaitFi can achieve a recognition accuracy of 94.2\% using two modalities, which significantly outperforms the state-of-the-art methods based on either WiFi or a camera. In the field of gait recognition, GaitFi innovatively proposes a feature-level fusion of visual modality and WiFi modality to compensate for the shortcomings of the two sensing modalities and achieve better performance.

The contributions of this paper are summarized as follows:
\begin{itemize}
    \item We study how vision and WiFi signals contribute to the human gait recognition task, and propose a multimodal human identification system, namely GaitFi. To the best of our knowledge, it is the first work for the WiFi-vision gait recognition method based on multimodal learning.
    \item In GaitFi, we propose a LRCN for WiFi and a boosted LRCN with LSTM for cameras, and then fuse them for deep metric learning. The fusion mechanism enables our system to leverage the complementarity of two modalities for better robustness.
    \item Real-world experiments demonstrate that our GaitFi outperforms the state-of-the-art gait recognition methods based on single WiFi or cameras.
\end{itemize}

The rest of the paper is organized as follows: \cref{sec:Related work} reviews WiFi and vision-based gait analytics. \cref{sec:Method} provides the detailed illustration of GaitFi. \cref{sec:Experiment} shows experiment procedure, results, comparison with existing works and ablation study. \cref{sec:Conclusion} concludes the paper and provides recommendations for future research topics.

\begin{figure*}[t]
  \centering
  \includegraphics[width=1\textwidth]{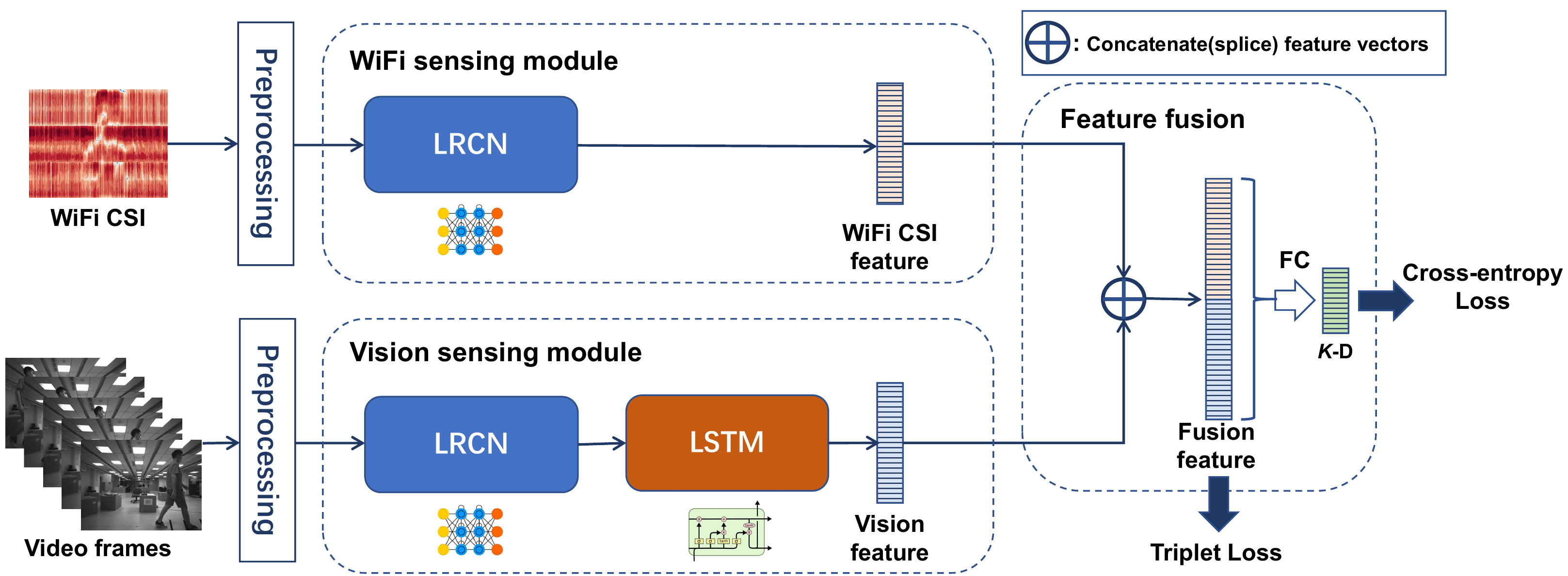}
  \caption{Structure and fusion mechanism of GaitFi system.}
  \label{fig: fusion}
\end{figure*}

\section{Related work}
\label{sec:Related work}

\subsection{WiFi-Based Sensing and Gait Recognition}

The WiFi-based gait recognition method uses RF signals from WiFi-enabled devices to determine human identity. The transmitter emits WiFi signals, which are reflected by different body parts of the walking subject and then recorded by CSI data at the receiver~\cite{yang2022benchmark}, which has empowered many applications including occupancy detection~\cite{zou2017freedetector}, crowd counting~\cite{zou2018device,zou2017freecount}, human activity recognition~\cite{zou2018deepsense,zou2017multiple,zou2019wifi,yang2018carefi,wang2021multimodal}, person identification~\cite{zou2018identification,wang2022caution}, vital sign detection~\cite{hu2022resfi}, pose estimation~\cite{yang2022metafi} and gesture recognition~\cite{zou2018robust,yang2019learning,zou2018gesture}. To use WiFi sensing in the real world, current research aims at efficient communication~\cite{yang2022efficientfi}, model security~\cite{yang2022robustsense} and data-efficient training~\cite{yang2022autofi}.

Recently, research on human identification using WiFi-enabled devices has begun to emerge because of ubiquitous WiFi-enabled IoT devices. This paper focuses on the research of WiFi in the field of gait recognition. A WiFi-based gait feature extraction system named WiFiU~\cite{wang2016gait} is proposed by Wang et al. to classify humans with different identities. Zhang et al.~\cite{zhang2016wifi} propose WiFi-ID, a WiFi-based gait recognition method that can be used in small offices or smart homes. Zeng et al.~\cite{zeng2016wiwho} utilize gait to recognize human identity by measuring the time domain information of WiFi signals. Lv et al.~\cite{lv2017wii} propose Wii, which improved gait recognition accuracy by performing autocorrelation on the torso reflection to remove imperfection in spectrograms. Cao et al.~\cite{cao2021lightweight} propose a lightweight deep learning algorithm named LW-WiID, which can achieve a relatively high recognition accuracy by extracting the spatial information of subcarriers. CAUTION~\cite{wang2022caution} proposes to employ few-shot learning for data-efficient human identification. From the above work, research on WiFi-based single-modal gait recognition is getting more appealing for IoT-enabled human identification.

\subsection{Vision-Based Gait Recognition}
Vision-based solution plays an essential role in gait recognition methods. Johansson et al. \cite{johansson1973visual} use moving light displays and reflectors on different joints of the human body and observe that gait patterns are unique, so that gait can be a biomarker feature that is recognized by vision. With the development of computer vision, vision-based gait recognition methods are gradually gaining widespread attention. Nowadays, vision-based gait recognition methods can be divided into two main categories, template-based and video sequence-based methods.

For the template-based gait recognition method, the gait silhouette contour sequence needs to be obtained using background subtraction \cite{wang2003silhouette}. Then, the resulting gait profile is aligned by cropping and then pixel-level operations are performed to generate a gait template, such as Gait Energy Image (GEI) \cite{han2005individual}. The obtained gait templates can be used to obtain feature representations by machine learning methods \cite{xing2016complete}. After obtaining the gait representation, the similarity between the representation pairs can be measured by metric learning methods \cite{takemura2017input}. Recently, an increasing number of deep learning methods have been applied to template-based gait recognition tasks \cite{wu2016comprehensive,he2018multi}. 

The video sequence-based gait recognition directly uses the silhouette sequence generated by background subtraction as the input to the deep learning neural network. This method can collect more temporal information, so specialized neural network structures need to be designed to extract such temporal information. Liao et al. \cite{liao2017pose} use a LSTM-based approach to extract temporal information from gait sequences. Chao et al. \cite{chao2019gaitset} propose GaitSet, a network that can blur time information of gait sequence. Lin et al. \cite{lin2021gait} propose to aggregate local temporal and local spatial information for gait recognition.
\subsection{Multimodal Machine Learning}

 Multimodal machine learning aims to build models that can process and correlate information from multiple modalities~\cite{ngiam2011multimodal}. The motivation for multimodal machine learning comes from the fact that every single modality has its own drawbacks that make them perform sub-optimally. In addition, humans perceive the world in a multimodal way, such as vision, sounds and text, encouraging the existence of multimodal learning. Therefore, when a research question or dataset contains multiple modalities, it is characterized as a multimodal task. Multimodal machine learning involves many research directions including representation, translation, alignment, fusion and co-learning. Fusion is responsible for combining the information of multiple modalities to perform target prediction (i.e., classification or regression). It is one of the earliest research directions of multimodal machine learning and is currently the most widely used one. According to the level of fusion, multimodal fusion can be divided into input-level~\cite{li2017pixel}, feature-level~\cite{ross2005feature,haghighat2016discriminant} and decision-level fusion~\cite{chatzis1999multimodal}. For our system GaitFi, the multimodal machine learning method is feature-level fusion.
Co-learning is another popular research topic in the multimodal machine learning domain, which can model resource-poor modalities by leveraging knowledge from other resource-rich modalities~\cite{rahate2022multimodal}. It achieves this capability by using transfer learning and domain adaptation methods~\cite{zou2019consensus}. There is also a type of work in co-learning called co-training~\cite{ning2021review}, which is responsible for studying how to expand a small number of annotations in multimodal data to obtain more annotation information.

\section{Method}
\label{sec:Method}

\subsection{WiFi-Vision Multimodal Gait Recognition Method}
\label{sec: System}

Different from the existing WiFi-based gait recognition methods that simply formulate the problem as a standard classification problem, we formulate it as a gait retrieval task that is more practical in reality. The gait retrieval task is similar to the visual pedestrian ReID task~\cite{lin2019improving} and visual gait recognition~\cite{chao2019gaitset}. Given gallery samples and probe samples (i.e. test samples), the objective is to find those samples in the probe that have the same identity as the gallery samples. Therefore, the process of gait recognition is to match the test gait sample with existing gallery gait data, which allows users to enlarge the categories easily in practice. Our system uses two modalities, WiFi and vision, to get richer gait information from different levels. As shown in \cref{fig:CSI example}, the two modalities can reflect the gaits of different people and indicate the occupancy condition. Then we introduce the two modalities of gait data.

\subsubsection{WiFi CSI Modality}
\label{sec: wifi csi}

\begin{figure*}
  \centering
  \begin{subfigure}{0.328\linewidth}
    \includegraphics[width=1\textwidth]{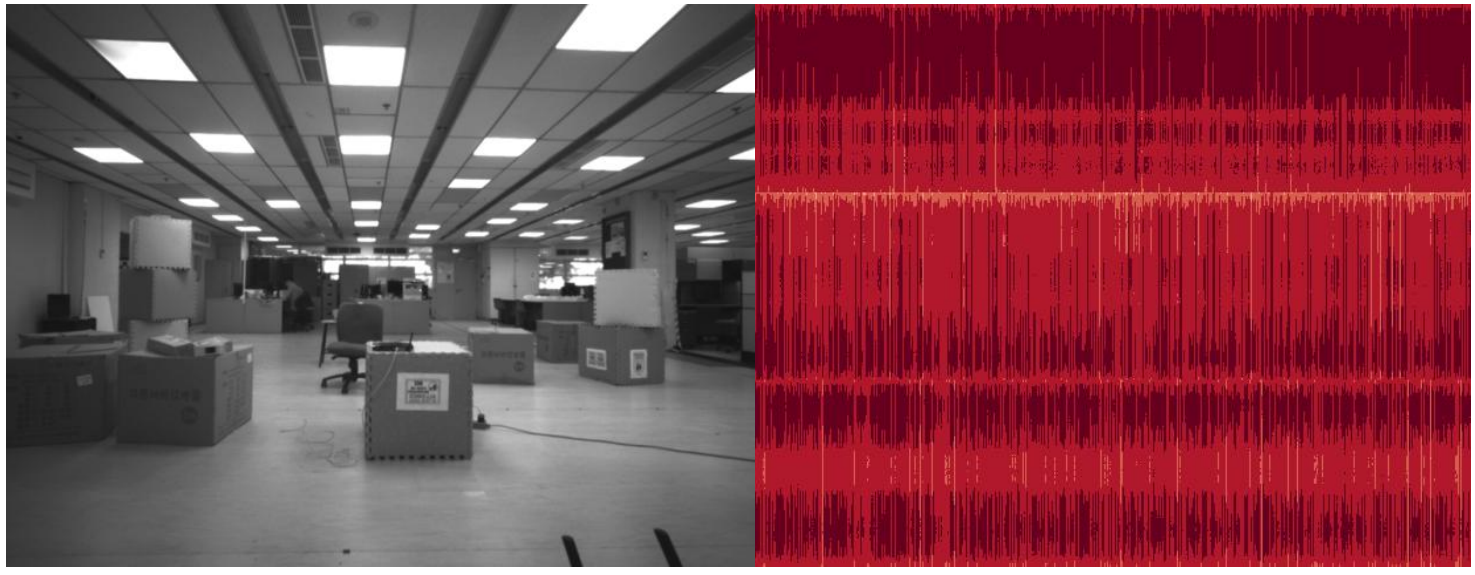}
    \caption{No subject.}
    \label{fig:No Subject}
  \end{subfigure}
  \hfill
  \begin{subfigure}{0.328\linewidth}
    \includegraphics[width=1\textwidth]{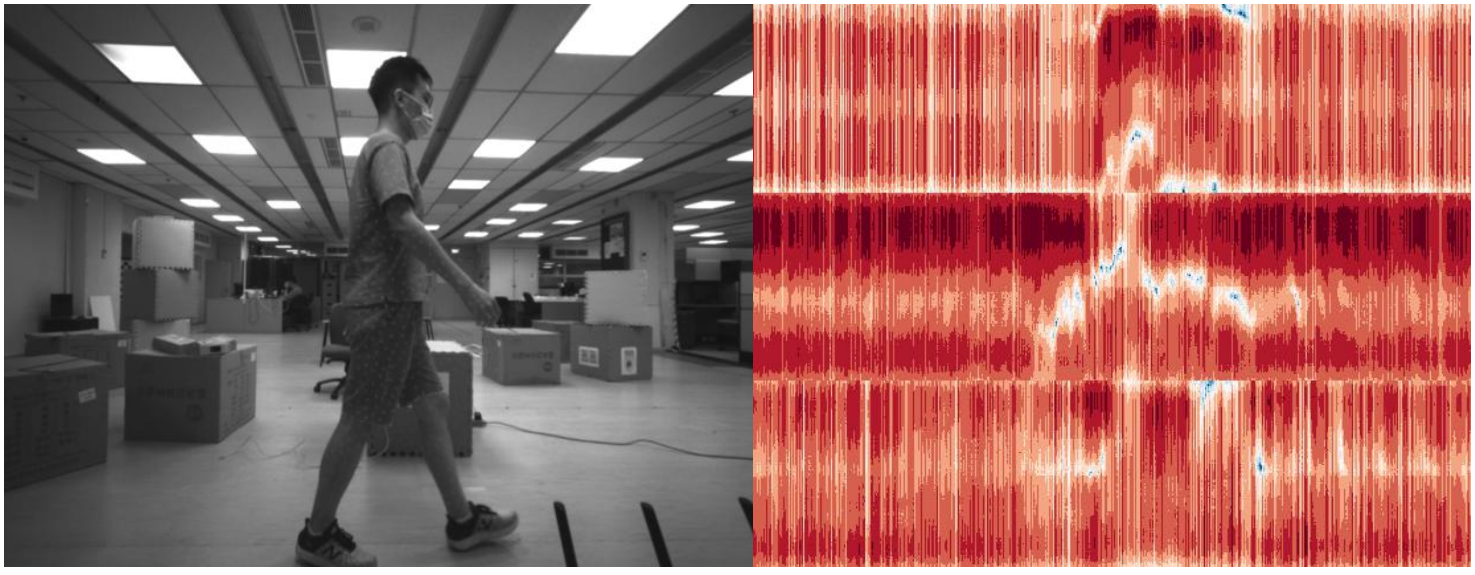}
    \caption{Subject A.}
    \label{fig:Subject A}
  \end{subfigure}
  \hfill
  \begin{subfigure}{0.328\linewidth}
    \includegraphics[width=1\textwidth]{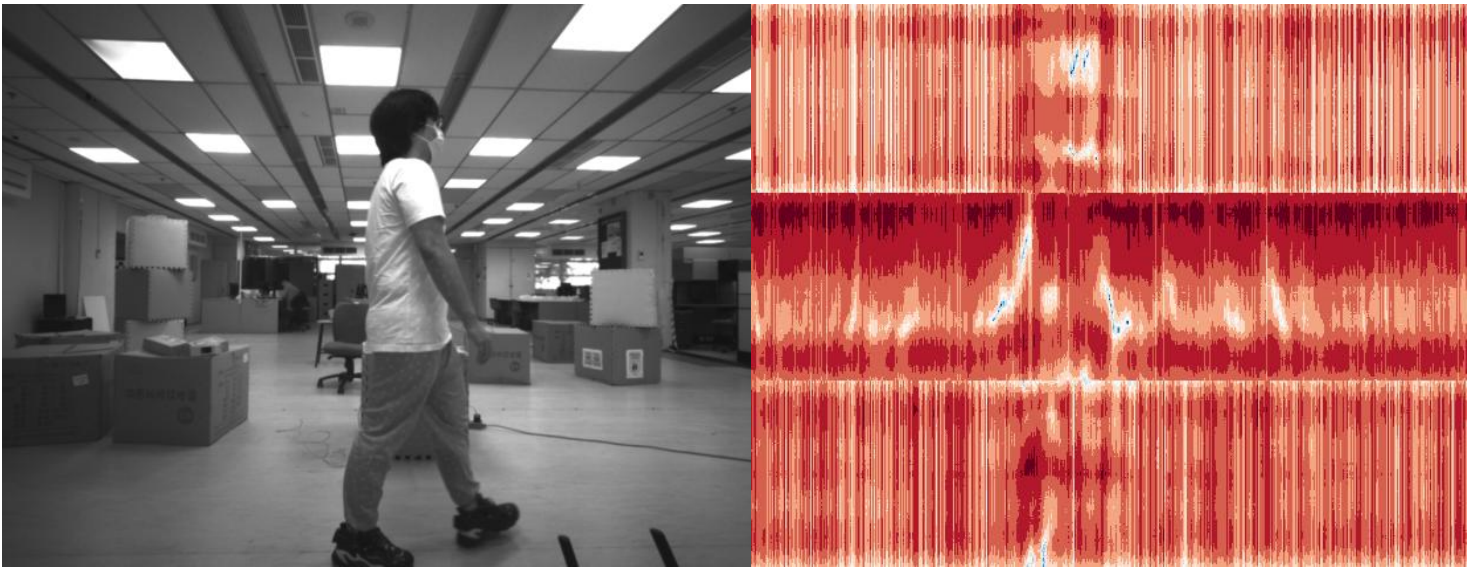}
    \caption{Subject B.}
    \label{fig:Subject B}
  \end{subfigure}
  \caption{Example of one image frame of a sample and the corresponding matrix (heatmap) of CSI data frame. Each row is one subcarrier and the column is the packet index (i.e., timestamp). The color denotes the amplitude value in the CSI matrix.}
  \label{fig:CSI example}
\end{figure*}

\begin{figure}
  \centering
  \includegraphics[width=1\linewidth]{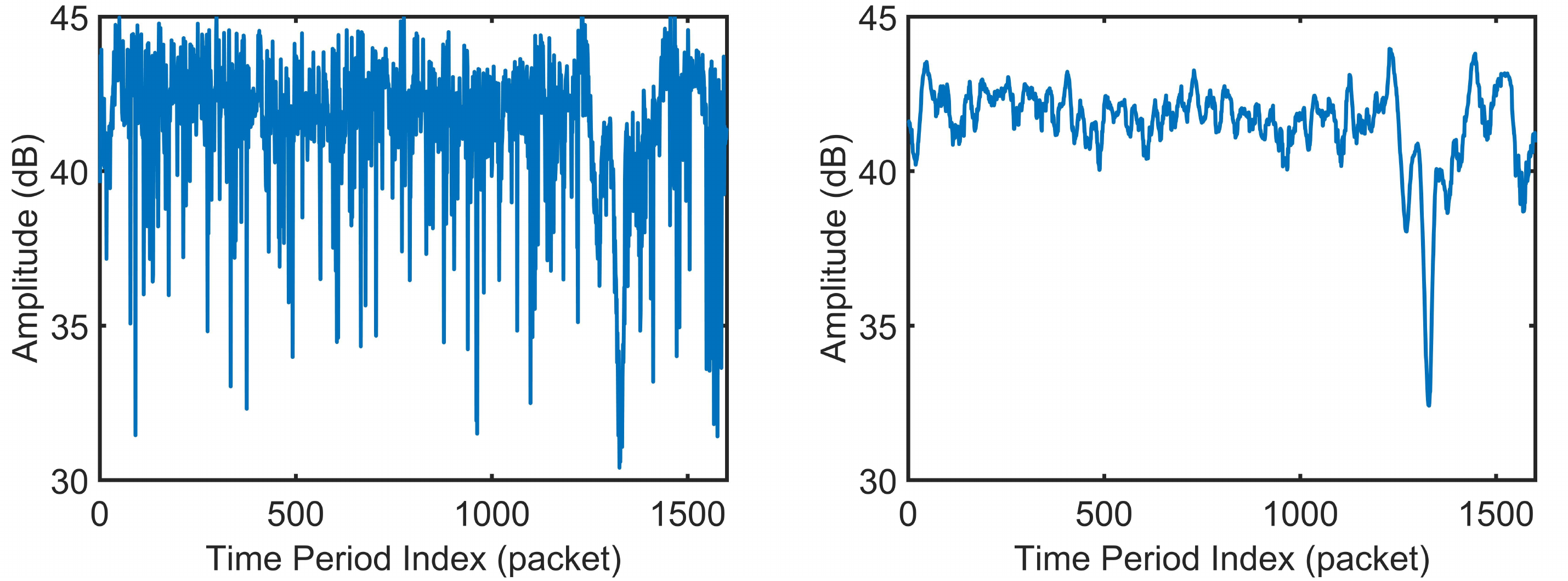}
  \caption{Comparison of raw and denoising data on the 50th subcarrier in the CSI
stream. (The left picture is the waveform of the subcarrier before denoising, and the right picture is the waveform of the subcarrier after denoising)}
  \label{fig:denoise}
\end{figure}

\begin{figure*}
  \centering
  \includegraphics[width=1\textwidth]{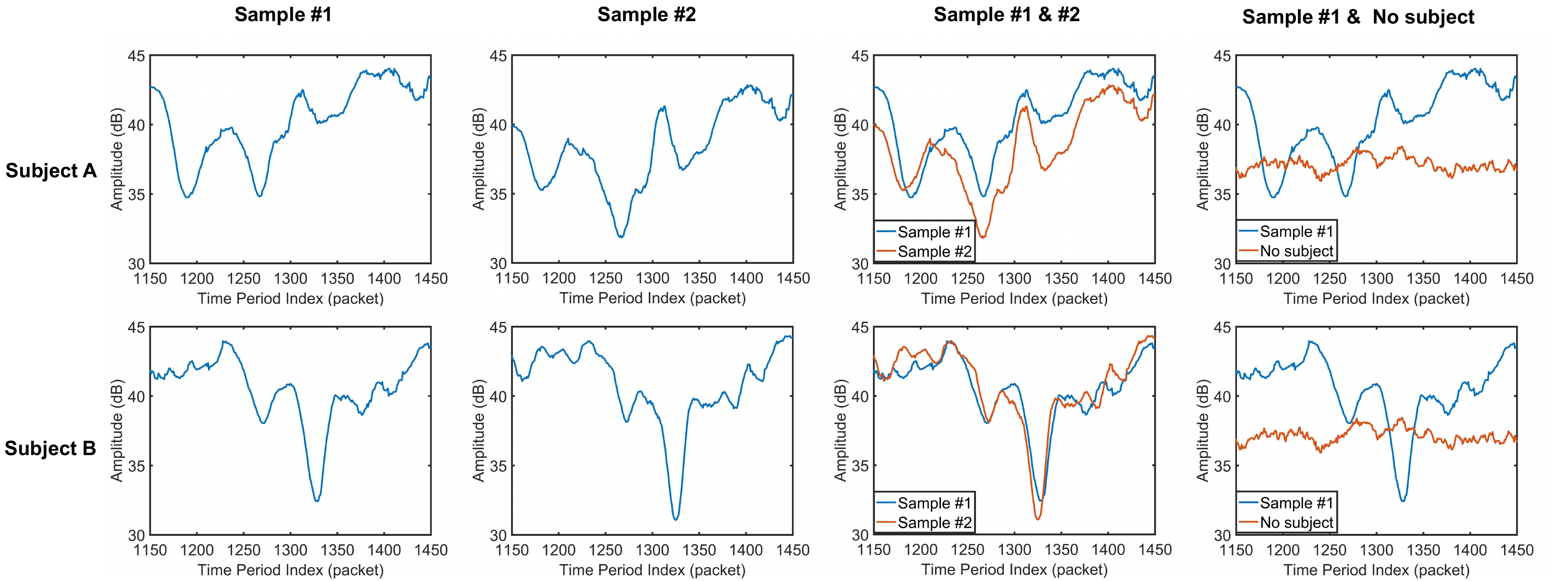}
  \caption{The CSI data of different subjects and vacant situation. }
  \label{fig: subcarrier}
\end{figure*}
WiFi signals transmit through multiple paths between the transmitter (TX) and the receiver (RX) of WiFi-enabled IoT devices, and these signals can be scattered and reflected by human motion between TX and RX~\cite{yang2018fine}. In wireless communication, the reflection, diffraction and scattering phenomena of WiFi signals affected by the physical environment can be described by channel state information (CSI)~\cite{yang2013rssi}. Modern WiFi devices use Orthogonal Frequency Division Multiplexing (OFDM) at the physical layer following the IEEE 802.11n/ac standard, which allows multiple transmit and receive antennas for Multiple-Input Multiple-Output (MIMO) communications. The CSI reveals fine-grained characterization of delay, amplitude decay, and multi-path phase-shift effects on each communication subcarrier~\cite{yang2018device}. We model the frequency domain of the WiFi signal as the channel impulse response $h(\tau)$
\begin{equation}
  h(\tau) = \sum_{m=1}^M\alpha_me^{j\phi_m}\delta(\tau-\tau_m),
  \label{eq:impulse}
\end{equation}
where $M$ denotes the total number of multipath, $\alpha_m$ and $\phi_m$ represent the amplitude and phase of the $m$-th multipath component, respectively, $\delta(\tau)$ denotes the Dirac delta function, and $\tau_m$ denotes time delay. However, due to limited WiFi bandwidth, only clusters of multipath components are distinguishable. In the frequency domain, a sampled version of the signal spectrum on each subcarrier can be obtained from RX, and the CSI measurements can be summarized as a complex number $H_i$
\begin{equation}
  H_i=|H_i|e^{j\angle H_i},
  \label{eq:Hi}
\end{equation}
where $|H_i|$ denotes the amplitude attenuation, and $\angle H_i$ denotes the phase shift at the $i$-th subcarrier. Due to the hardware and the environmental variations, the carrier frequency drifts, and the robustness of the phase information is relatively poor~\cite{gjengset2014phaser}. Thus, we only use the amplitude information in our system. We employ two TP-Link N750 routers with a modified OpenWrt firmware to collect CSI data~\cite{yang2018device}. The modified firmware is equipped with the Atheros CSI tool~\cite{xie2015precise} that enables routers to record the packets transmitted over the wireless channel and extract CSI measurements from those packets. The routers are set to run in a 40MHz channel when operating at 5GHz, which allows us to extract 114 subcarriers of CSI for each TX-RX pair. At each measurement, the number of CSI streams we can obtain is $N_{total}$
\begin{equation}
  N_{total}=N_{TX}N_{RX}N_{subcarriers},
  \label{eq:number}
\end{equation}
where $N_{TX}$ and $N_{RX}$ denote the number of antennas of the router that transmits the signal and the router that receives the signal, respectively, and $N_{subcarriers}$ denotes the number of subcarriers that is 114 in a 40MHz channel. For the gait recognition task, the CSI data frames are generated when subjects walk through the Line-of-Sight (LoS) path of WiFi signal propagation (between TX and RX). The gait is unique for a subject as illustrated in \cref{fig:CSI example} with 3 examples, where we use the heatmap to visualize the CSI data in different situations. \cref{fig:No Subject} shows the heatmap of the CSI data when no subject passes the experimental site. \cref{fig:Subject A} and \cref{fig:Subject B} show the CSI data frames when subjects \textit{A} and \textit{B} pass by, respectively. It can be seen intuitively from the \cref{fig:CSI example} that the CSI pattern is unique for different subjects. Therefore, the CSI data extracted from off-the-shelf WiFi routers can be used for gait recognition.

In order to analyze the effect of gait on WiFi CSI, we visualize one of the CSI subcarriers (the 50th out of 114) for analysis. For better resolution, we use the moving average method~\cite{isufi2016autoregressive} for denoising, as shown in \cref{fig:denoise}, where the y-axis is the amplitude attenuation represented by $|H_i|$ in \cref{eq:Hi} and the x-axis is the packet number which can also be expressed in terms of the length of the received packets. In order to illustrate the correlation between gait and WiFi CSI data, we select two WiFi CSI samples of two subjects and visualize the 50th subcarrier in \cref{fig: subcarrier}. It is observed that the presence of subjects leads to obvious CSI variations, and the CSI patterns of the same subject are similar, which illustrates that the CSI can reflect the unique human gait biomarker. This phenomenon provides a factual basis for using WiFi modality to recognize human gait.

\subsubsection{Vision Modality}
\label{sec: camera}

Computer vision has been applied to many tasks, such as object detection~\cite{redmon2016you} and human activity recognition~\cite{ahmad2021human}. Recently, vision-based gait recognition has been extensively studied and has achieved remarkable accuracy due to the wide utilization of cameras and the development of computer vision~\cite{alzubaidi2021review}. In our system, we use a camera to obtain data on vision modality. The camera is set close to the WiFi receiver to capture video of subjects' gaits simultaneously. Each video sample consists of a series of frames that contain the continuous temporal data of human gaits. We synchronize the CSI and video data in our system for better multimodal fusion.

\subsection{Multimodal Learning for WiFi and Vision}
\label{sec: multimodal Gait Recognition Model}


Having data from the two modalities, the multimodal learning module accounts for representing and recognizing human gaits. Given a set of $N$ subjects with $M$ samples per subject, we denote the dataset as $D^T$
\begin{equation}
  D^T=\{(x^{ij}_w,x^{ij}_v),y^{ij}\},i\in [1,N],j\in [1,M],
  \label{eq:DT}
\end{equation}
where $x^{ij}_w$ denotes the $j$-th sample of the $i$-th subject in the WiFi modality, $x^{ij}_v$ denotes the corresponding sample of vision modality and $y^{ij}$ denotes the ground truth of its human ID.
The objective of our GaitFi is to map a test sample to its subject ID, denoted as
\begin{equation}
  y^{ij}=\Phi(x^{ij}_w,x^{ij}_v),
  \label{eq:map}
\end{equation}
where $\Phi(\cdot)$ is our gait recognition model.

\subsubsection{WiFi Gait Recognition Module}

\begin{table}[ht]
\renewcommand\arraystretch{1.5}
\setlength\tabcolsep{10pt}
\centering
\caption{Structure of LRCN and WiFi-LRCN. LRCN: A network for extracting features when two modalities are fused. WiFi-LRCN: Optimized LRCN specifically for WiFi modality. ($Res$ denotes the residual block proposed by He et al.~\cite{He_2016_CVPR}, and $Ch$ represents the channel)}
\label{tab: GaitFi-Res}
\scalebox{0.7}{
\begin{tabular}{|c|c|c|}
\toprule
\multicolumn{1}{|l|}{\textbf{Block name}}    & \multicolumn{1}{c|}{\textbf{LRCN}}              & \textbf{WiFi-LRCN (only WiFi)} \\ \midrule\midrule
\multicolumn{1}{|l|}{Conv\_1}       & \multicolumn{1}{c|}{3x3 conv, stride 2, $Ch$ 8}    & 7x21 conv, stride 5, $Ch$ 64           \\ \midrule
\multicolumn{1}{|l|}{Res-layer\_1}  & \multicolumn{1}{c|}{$Res\left(
    \begin{array}{c}
        3\times3,\quad Ch=8\\
        3\times3,\quad Ch=8\\
    \end{array}
\right)\times2$}            & $Res\left(
    \begin{array}{c}
        3\times3,\quad Ch=64\\
        3\times3,\quad Ch=64\\
    \end{array}
\right)\times2$                \\ \midrule
\multicolumn{1}{|l|}{Conv\_2}       & \multicolumn{1}{c|}{3x3 conv, stride 2, $Ch$ 16}   & 3x7 conv, stride 1, $Ch$ 64            \\ \midrule
\multicolumn{1}{|l|}{Res-layer\_2}  & \multicolumn{1}{c|}{$Res\left(
    \begin{array}{c}
        3\times3,\quad Ch=16\\
        3\times3,\quad Ch=16\\
    \end{array}
\right)\times2$}            & $Res\left(
    \begin{array}{c}
        3\times3,\quad Ch=128\\
        3\times3,\quad Ch=128\\
    \end{array}
\right)\times2$               \\ \midrule
\multicolumn{1}{|l|}{Pool\_1}       & \multicolumn{1}{c|}{\textbackslash{}}         & MaxPool2d, kernel=1x2, stride=1x2 \\ \midrule
\multicolumn{1}{|l|}{Conv\_3}       & \multicolumn{1}{c|}{3x3 conv, stride 2, $Ch$ 32} & 3x7 conv, stride 1, $Ch$ 256           \\ \midrule
\multicolumn{1}{|l|}{Pool\_2}       & \multicolumn{1}{c|}{\textbackslash{}}         & MaxPool2d, kernel=1x2, stride=1x2 \\ \midrule
\multicolumn{1}{|l|}{Res-layer\_3}  & \multicolumn{1}{c|}{$Res\left(
    \begin{array}{c}
        3\times3,\quad Ch=32\\
        3\times3,\quad Ch=32\\
    \end{array}
\right)\times2$}            & $Res\left(
    \begin{array}{c}
        3\times3,\quad Ch=512\\
        3\times3,\quad Ch=512\\
    \end{array}
\right)\times2$              \\ \midrule
\multicolumn{1}{|l|}{FC\_1}         & \multicolumn{1}{c|}{Linear, 64}               & Linear, 512                       \\ \midrule
\multicolumn{1}{|l|}{FC\_2}          & \multicolumn{1}{c|}{\textbackslash{}}         & Linear, 12                        \\ \bottomrule 
\end{tabular}}

\end{table}

In order to perform gait recognition based on WiFi data $x^{ij}_w$, we need to extract spatial features across subcarriers and temporal features across time~\cite{sheng2020deep}. To this end, we do not use the prevailing models in the computer vision field, and propose a Lightweight Residual Convolution Network (LRCN), where the main blocks and related parameters are shown in \cref{tab: GaitFi-Res}. Considering the efficiency, we decrease the model complexity while preserving the capacity. The LRCN includes convolution blocks, residual blocks, batch normalization layers, ReLU layers, a flatten layer and a fully connected layer. The CSI frame input to the network first passes through a convolution block with a kernel of $3\times3$ and a stride of $2\times2$ with the number of channels to 8. Then a batch normalization layer and a ReLU layer are used for better convergence. To extract features by deeper layers, we design 3 convolution blocks with 8, 16 and 32 channels, each followed by a residual block. Compared to classic ResNet-18, our design has smaller parameters and floating-point operations (FLOPs).

In the LRCN, the convolution layers and residual blocks are to extract part of the features of the gait, where the residual block in the LRCN consists of two residual blocks proposed by He et al.~\cite{He_2016_CVPR}. The mathematical formula of the residual module can be expressed as
\begin{equation}
  \mathcal{F}(x)=\mathcal{C}(x)+x,
  \label{eq:res}
\end{equation}
where $x$ denotes the input feature, $\mathcal{F}(\cdot)$ is the residual block and $\mathcal{C}(\cdot)$ is the convolution layers. The residual design mitigates the degradation problem of deep neural networks when we increase the depth of deep neural networks. Batch normalization layers further address the problem of gradient vanishing and help attain better performance~\cite{ioffe2015batch}. The ReLU layer as an activation layer is to add non-linearity for better model capacity. As the CSI patterns are complicated and non-linear, we use all these layers in the LRCN. After feature extraction, we flatten the features into a 64-dimensional feature space. We denote this 64-dimensional feature as $z^{ij}_w$
\begin{equation}
  z^{ij}_w=F^{\theta_w}_w(x^{ij}_w),
  \label{eq:zw}
\end{equation}
where $F_w(\cdot)$ denotes the forward functions of the LRCN model, parameterized by $\theta_w$. In \cref{tab: GaitFi-Res}, we also propose a WiFi-LRCN network that leverages more parameters, which is only for comparison in the experiments. The WiFi-LRCN can achieve better performance in the single modality situation, but cannot lead to further improvement for multimodal performance with an increase in complexity of the algorithm.

\begin{algorithm}[t]
 \small
 \SetAlgoLined
 \SetAlgoLongEnd
 \DontPrintSemicolon
 \SetKwInput{KwModule}{Module}
 \SetKw{KwBegin}{BEGIN:}
 \SetKw{KwEnd}{END.}
 
 \small
 \caption{The algorithm of GaitFi system. \label{algo:System}}
 
 \textbf{Step 1: Training Phase}\;
 \KwModule{
  the LRCN for wifi modality $F^{\theta_w}_{w}$, the LRCN for vision modality $F^{\theta_v}_{w}$, the LSTM $G_l^{\theta_l}$, the fusing function $\Psi(\cdot,\cdot)$, the mapping function $L$
 }
 \KwIn{labeled samples $\{(x^{ij}_w,x^{ij}_v),y^{ij}\}_{i=1,j=1}^{N,M}$}
 
 \KwBegin{}\;
 
 \While(){epoch $<$ total epoch}{
  
  Obtain the fusing feature via $z^{ij}_u=\Psi(F^{\theta_w}_{w}(x^{ij}_w),G_l^{\theta_l}(F^{\theta_v}_{w}(x^{ij}_v)))$

  Map to another feature space via $z^{ij}_r=L(z^{ij}_u)$
  
  Calculate $\mathcal{L}_{triplet}$ by $z^{ij}_u$ 
  
  Calculate $\mathcal{L}_{ce}$ by $z^{ij}_r$
  
  Update $\theta_w,\theta_v,\theta_l$ by minimizing $\mathcal{L}_{ce}+\alpha \mathcal{L}_{triplet}$
 }
 
 \KwOut{the model parameters $\theta_w,\theta_v,\theta_l$}
 \KwEnd{}
 
 ~\\
 
 \textbf{Step 2: Testing Phase}\;
 
 \KwIn{an unlabeled sample in the probe $(x_w,x_v)$, labeled samples in the gallery $\{(x^{ij}_w,x^{ij}_v),y^{ij}\}_{i=1,j=1}^{N,M}$}
 
 \KwBegin{}\;
 
 Obtain the fusing feature vectors: $z_u=\Psi(F^{\theta_w}_{w}(x_w),G_l^{\theta_l}(F^{\theta_v}_{v}(x_v)))$

 \While(){$i\in [1,N]$, $j\in [1,M]$}{
 
 $z^{ij}_u=\Psi(F^{\theta_w}_{w}(x^{ij}_w),G_l^{\theta_l}(F^{\theta_v}_{v}(x^{ij}_v)))$
 }
 
  \While(){$i\in [1,N]$}{
 $d^i=\sum^{M}_{j=1}||z_u-z^{ij}_u||^2$
 }
 
 $y\leftarrow \arg \min_{d^i} y^i$

 \KwOut{the label $y$ of the testing sample}
 \KwEnd
\end{algorithm}

\subsubsection{Visual Gait Recognition Module}
The video data is composed of a sequence of consecutive gait image frames, denoted as $x^{ij}_v$ in \cref{eq:DT}. As the video sequence consists of much temporal information, we further utilize the Long Short-Term Memory network (LSTM) after the LRCN. Specifically, we firstly use the LRCN to extract the frame-level features, and get 64-dimensional features for each frame. Since the LSTM better captures the dependencies of consecutive frames, we input the frame-level features in chronological order into a LSTM with 64 hidden states, generating the video features that are denoted as $z^{ij}_v\in \mathbb{R}^{64}$
\begin{equation}
  z^{ij}_v=G_l^{\theta_l}(F^{\theta_v}_v(x^{ij}_v)),
  \label{eq:zV}
\end{equation}
where $F_v(\cdot)$ denotes the LRCN model acting on vision modality, parameterized by $\theta_v$, and $G_l(\cdot)$ denotes the LSTM model, parameterized by $\theta_l$.

\subsubsection{Modality Fusion and Learning Objectives}
\label{sec: Modal Fusion and Learning Objectives}

After extracting the WiFi CSI feature vector $z^{ij}_w$ and image sequence feature vector $z^{ij}_v$ through the WiFi module and vision module, we propose a modality fusion mechanism to fuse two modalities. \cref{fig: fusion} shows the whole process from feature extraction to modal fusion. We concatenate the two feature vectors into a multimodal feature vector. In this way, we associate the feature information of the two modalities and obtain a higher-dimensional and more discriminative feature space. After this, we use a fully connected layer and \textit{softmax} function to map the multimodal feature to a $K$-dimensional feature, where $K$ denotes the number of subjects. The multimodal feature vector and the $K$-dimensional feature vector are denoted as $z^{ij}_u$ and $z^{ij}_r$, respectively. The whole process of modality fusion can be formulated as
\begin{equation}
\label{eq}
\left\{
\begin{aligned}
z^{ij}_u&=\Psi(z^{ij}_w,z^{ij}_v) \\
z^{ij}_r&=L(z^{ij}_u),
\end{aligned}
\right.
\end{equation} 
where $\Psi(\cdot,\cdot)$ denotes the operation of feature concatenation, and $L(\cdot)$ denotes the linear mapping operation for a fully connected layer.
After obtaining the multimodal feature $z^{ij}_u$ and the $z^{ij}_r$ through modality fusion, we design the objectives to train our model in an end-to-end manner. For $z^{ij}_u$, we aim to obtain a metric feature space where the samples from the same subject can cluster, so we use the triplet loss to implement similarity calculation between samples~\cite{hermans2017defense}. The triplet loss can pull the samples of the same category close while pushing those of different categories away, which is formulated as $\mathcal{L}_{triplet}$
\begin{equation}
  \begin{aligned}
  \mathcal{L}_{triplet}=\max(||z^{ij}_u-z^{im}_u||^2-||z^{ij}_u-z^{pn}_u||^2+\eta,0), \\ 
  \end{aligned}
  \label{eq:triplet}
\end{equation}
where $i\neq p$, $j\neq m\neq n$, $\max(\cdot, \cdot)$ denotes the function of taking the maximum value, $z^{ij}_u,z^{im}_u,z^{pn}_u$ are the multimodal features of three samples, and $\eta$ denotes the manually set margin, which is set to 0.2 empirically. The second objective is the normal cross-entropy loss for a $K$-way classification. To this end, we obtain the softmax outputs of $z^{ij}_r$ in each dimension $k$, denoted as $s^{ij}_{(k)}$
\begin{equation}
  s^{ij}_{(k)}=\frac{\exp(z^{ij}_{r(k)})}{\Sigma^{K}_{c=1}\exp(z^{ij}_{r(c)})},k\in [1,K],
  \label{eq:softmax}
\end{equation}
where $\exp(\cdot)$ denotes the exponential function, $z^{ij}_{r(k)}$ is the value of the $k$-th dimension of the $K$-dimensional feature for the $j$-th sample of the $i$-th class. Then we calculate cross-entropy loss~\cite{shore1981properties}:
\begin{equation}
  \mathcal{L}_{ce}=-\Sigma^{K}_{k=1}y^{ij}_{o(k)}\log s^{ij}_{(k)},
  \label{eq:cross-entropy}
\end{equation}
where $y^{ij}_{o(k)}$ is the value of the $k$-th dimension of the one-hot label $y^{ij}_o$. The final objective $\mathcal{L}_{total}$ is written by
\begin{equation}
  \mathcal{L}_{total}=\mathcal{L}_{ce}+\alpha \mathcal{L}_{triplet},
  \label{eq:loss}
\end{equation}
$\alpha$ controls the ratio of the metric learning loss in the total loss. The loss can be optimized via backpropagation by updating $\theta_w$, $\theta_v$ and $\theta_l$. The testing process is based on metric measurement by finding the most similar subject cluster via Euclidean distance. The training and testing of our system are summarized in \cref{algo:System}.

\begin{figure*}
    \centering
    \begin{subfigure}{0.27\linewidth}
        \includegraphics[width=1\textwidth]{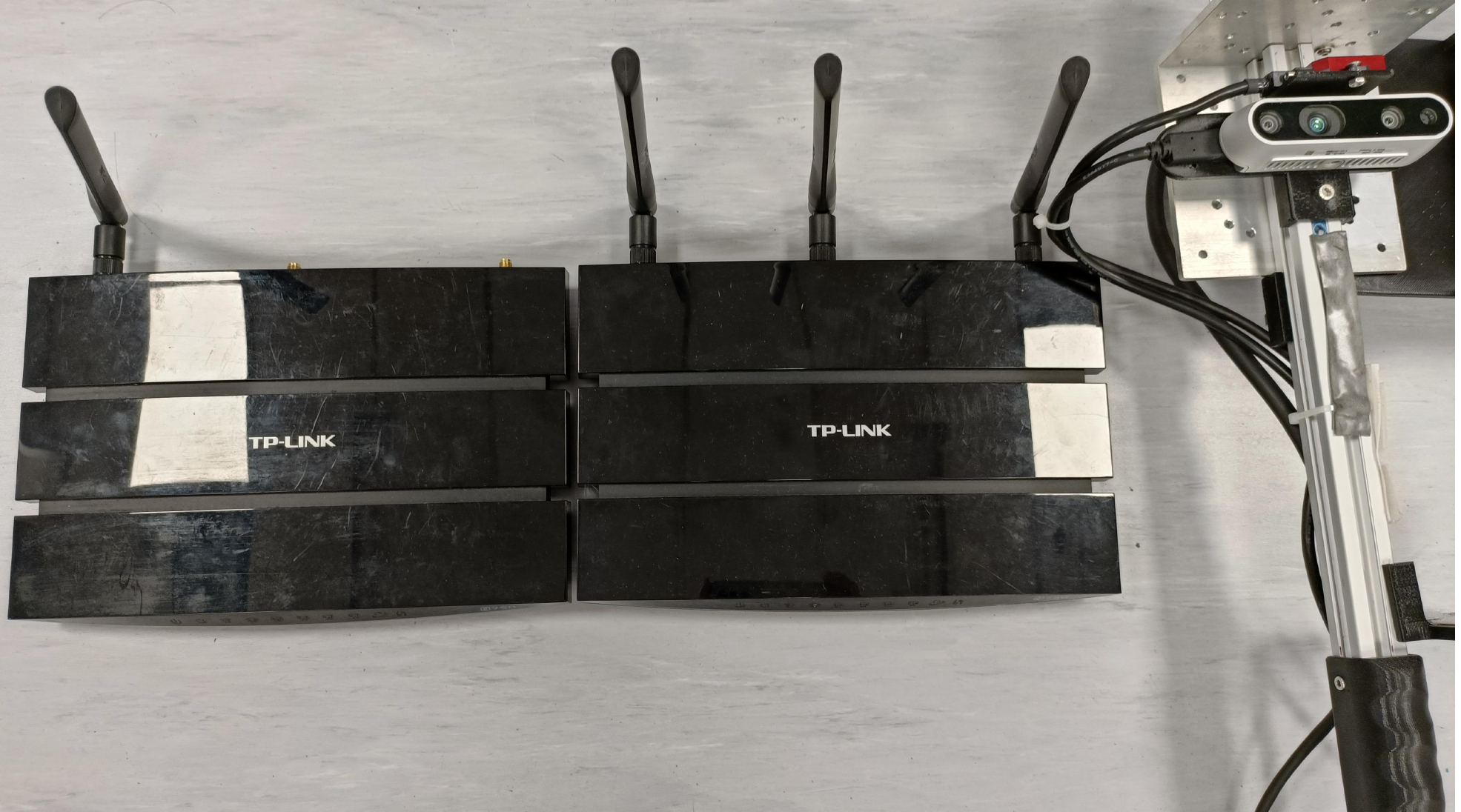}
        \caption{The routers (TP-Link N750) and the camera (Intel RealSense).}
        \label{fig:sensor}
    \end{subfigure}
    \hfill
    \begin{subfigure}{0.35\linewidth}
        \includegraphics[width=1\textwidth]{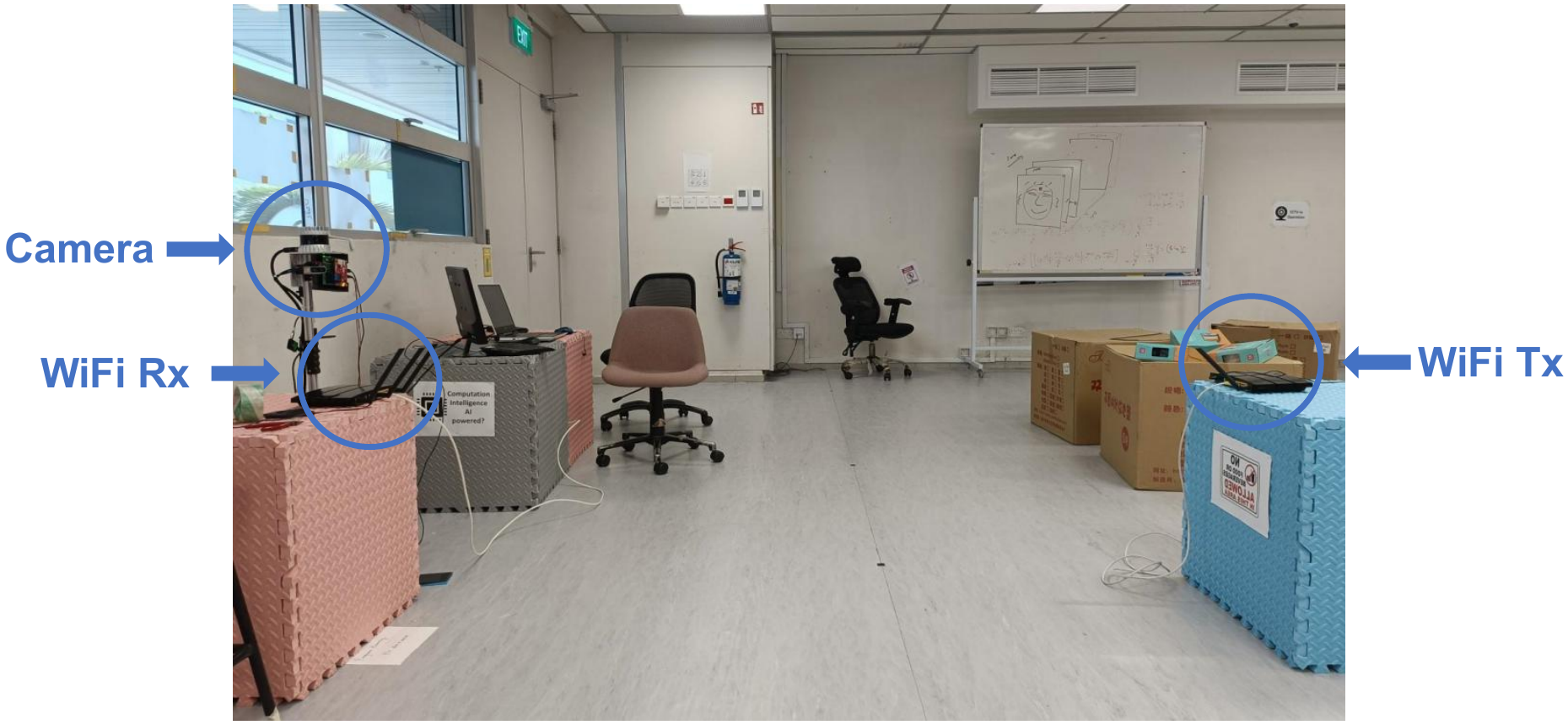}
        \caption{RGB photo of the testbed.}
        \label{fig:plat}
    \end{subfigure}
    \hfill
    \begin{subfigure}{0.35\linewidth}
        \includegraphics[width=1\textwidth]{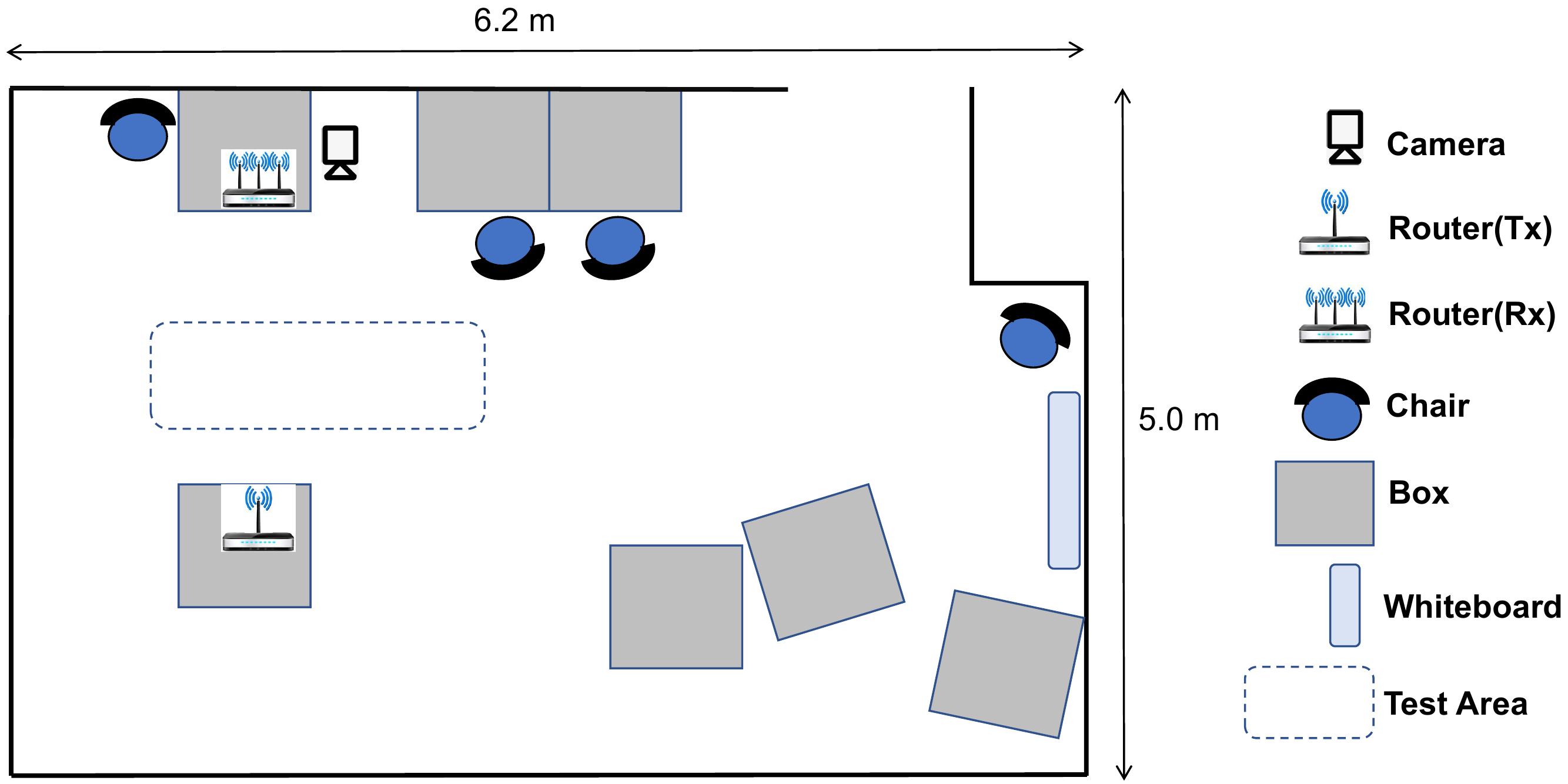}
        \caption{The layout of the testbed.}
        \label{fig:layout}
    \end{subfigure}
    \caption{The real scenes and the layout of the testbed for our method.}
    \label{fig:testbed}
\end{figure*}


\begin{figure}
  \centering
  \includegraphics[width=1\linewidth]{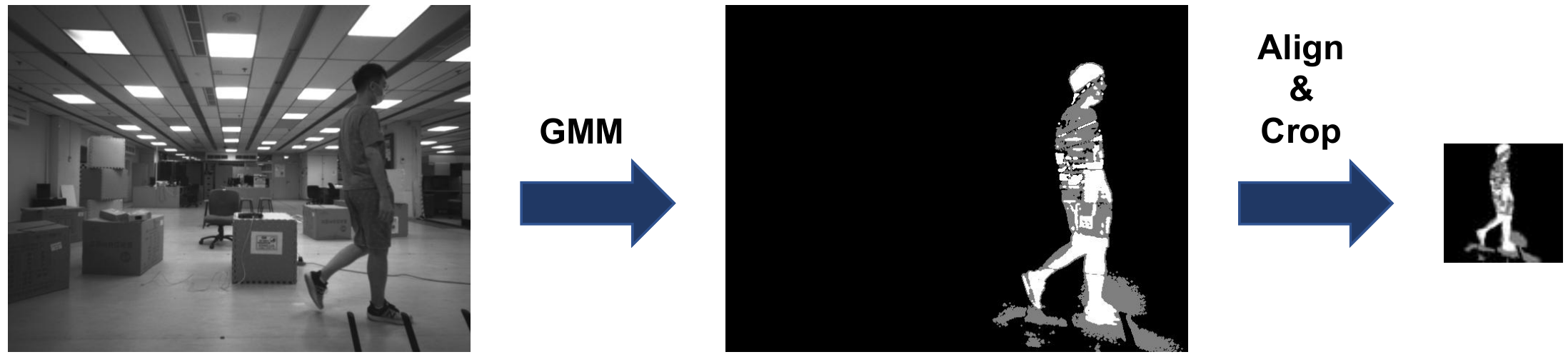}
  \caption{Vision data frame preprocessing: first use GMM to get silhouettes, then align and crop.}
  \label{fig:silhouette}
\end{figure}

\section{Experiment}
\label{sec:Experiment}

\subsection{Setup and Data Collection}
\label{sec:Testbed}

\textbf{System setup.} To evaluate the performance of the GaitFi system using the gait recognition method, we use two commercial TP-LINK N750 routers as WiFi transmitters and receivers respectively to acquire CSI data, and an Intel RealSense camera to acquire vision data for human gaits. The routers and the camera are shown in \cref{fig:sensor}. The testbed is set up in an indoor environment, as shown in \cref{fig:plat}, where the camera and the router as the WiFi receiver are on one side of the photo, while the router as the WiFi transmitter is set on the other side. The receiver and the RealSense camera are connected to the same mini-PC for synchronization and data annotations. The WiFi routers are set to run at 5GHz with 40MHz bandwidth, whose firmware is upgraded as described in \cref{sec: wifi csi} to collect 114 subcarriers of CSI data for each TX-RX pair. The receiver is equipped with 3 antennas while the transmitter is equipped with 1 antenna. The distance of the TX-RX pair is $2.1 m$. \cref{fig:layout} is a top view of the testbed layout, where the sensing devices and the facilities are illustrated.

\textbf{Data collection.} To test the performance of the GaitFi system, we collect a dataset for performance evaluation on the above platform. We invite 12 volunteers with heights between $1.55 m$ and $1.85 m$ as subjects, where their genders and heights are shown in \cref{tab: dataset}. In comparative experiments, this dataset can effectively demonstrate the advantage and correctness of using GaitFi, compared to other gait recognition methods. The area of the dashed box in \cref{fig:layout} shows the area where the subjects walk, while the walking direction of each subject is perpendicular to the line of sight (LoS) of the two routers. Walking from one side to the other side is recorded as a sample. 30 samples (i.e., 15 back and forth) are collected for each subject. The WiFi sensor and camera sensor simultaneously record $2s$ of gait information for each walk. In this manner, we can obtain gait samples of 12 different groups (i.e., subjects), and each group contains 30 WiFi CSI frames and corresponding gait videos. During training, 20 samples of each subject serve as the training set, i.e., the gallery set, while the remaining 10 samples are utilized as the probe set. The gait videos obtained by the camera are drawn at an interval of $0.035 s$ to form the original visual gait frame sequence, and the pixel size of each frame is $640\times480$. For the CSI data obtained by the WiFi sensor, the sampling rate of the receiver is 800 $packets/s$, and the sensing time is $2s$, so a WiFi CSI data frame has 1600 $packets$. Because the transmitter router has 1 antenna, and the receiver router has 3 antennas, with \cref{eq:number}, the size of each WiFi CSI data frame is $3\times114\times1600$. 

\begin{table}[ht]
\centering
\caption{Statistics of subjects in our dataset.}
\label{tab: dataset}
        \begin{tabular}{c|c|c}
            \toprule
            Identity label & Gender & Height ($m$)  \\
            \midrule
            \#1 & Male & 1.80\\
            \#2 & Male & 1.74\\
            \#3 & Male & 1.78\\
            \#4 & Male & 1.77\\
            \#5 & Female & 1.58\\ 
            \#6 & Female & 1.68\\
            \#7 & Female & 1.70\\
            \#8 & Female & 1.63\\
            \#9 & Male & 1.70\\
            \#10 & Male & 1.78\\
            \#11 & Male & 1.75\\
            \#12 & Female & 1.69\\
            \bottomrule
        \end{tabular}
\end{table}

\begin{figure*}
  \centering
  \begin{subfigure}{0.328\linewidth}
    \includegraphics[width=1\linewidth]{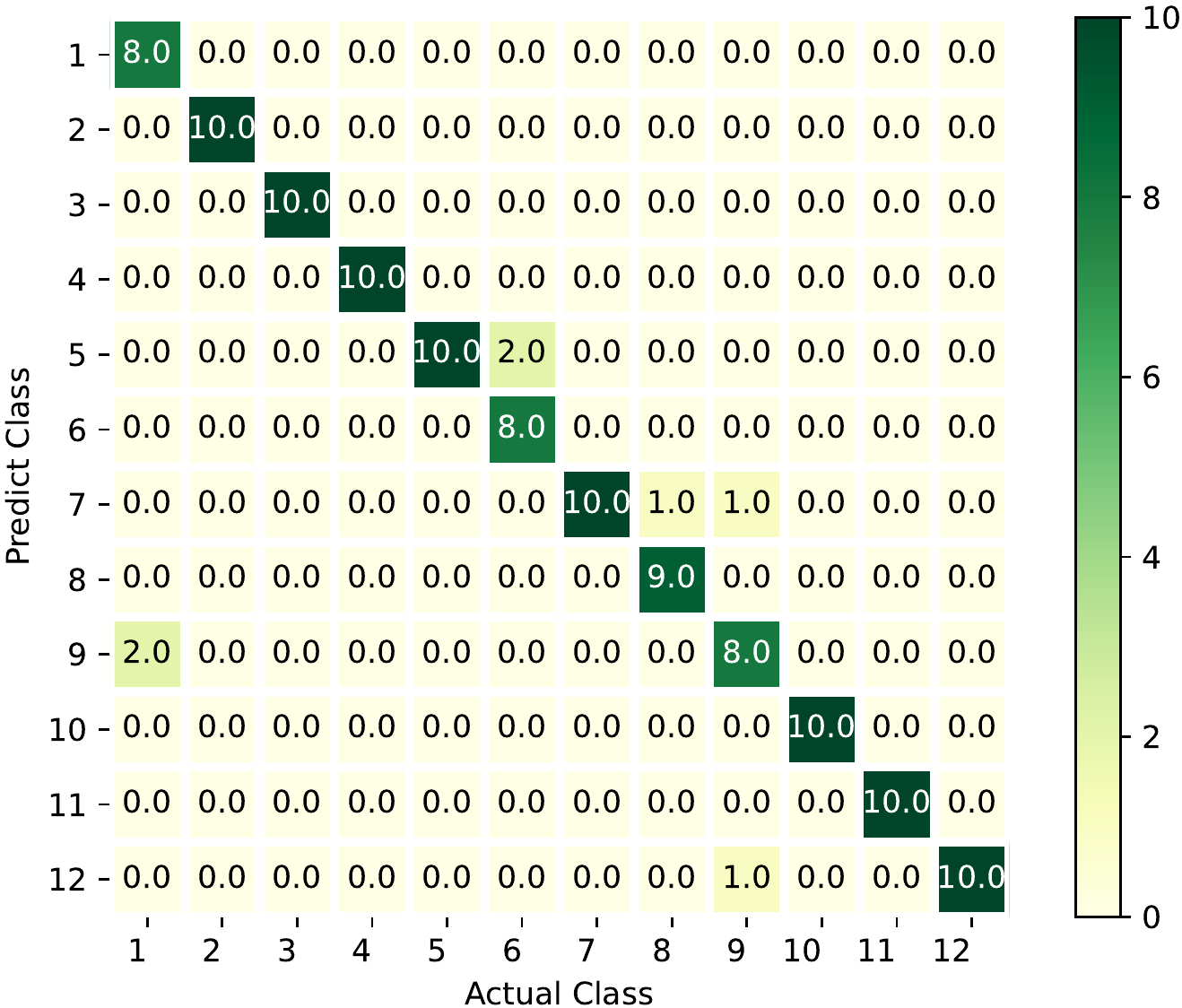}
    \caption{Confusion matrix of the GaitFi.}
    \label{fig:fusion matrix}
  \end{subfigure}
  \hfill
  \begin{subfigure}{0.328\linewidth}
    \includegraphics[width=1\linewidth]{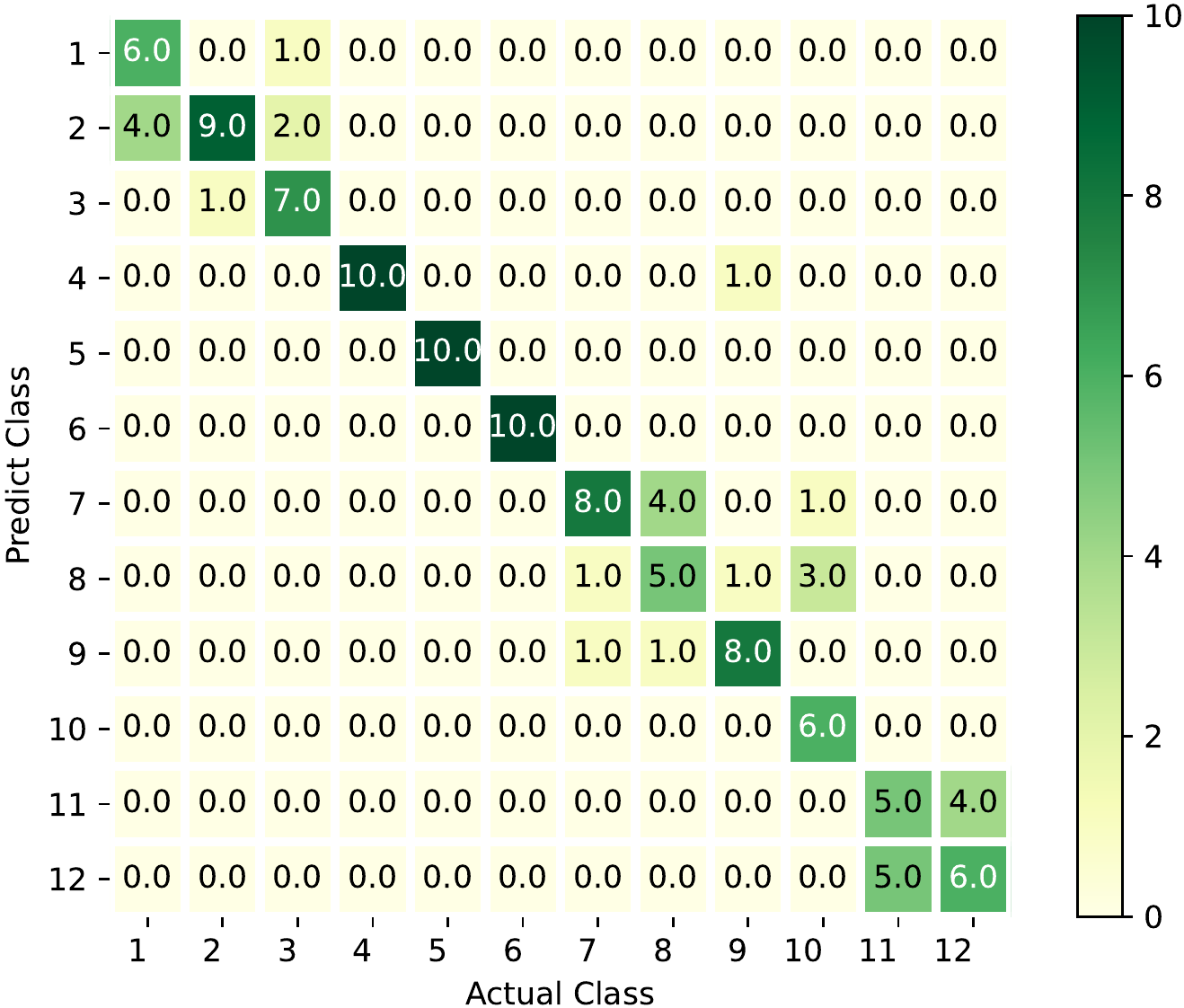}
    \caption{Confusion matrix of the WiFi modality.}
    \label{fig:wifi matrix}
  \end{subfigure}
  \hfill
  \begin{subfigure}{0.328\linewidth}
    \includegraphics[width=1\linewidth]{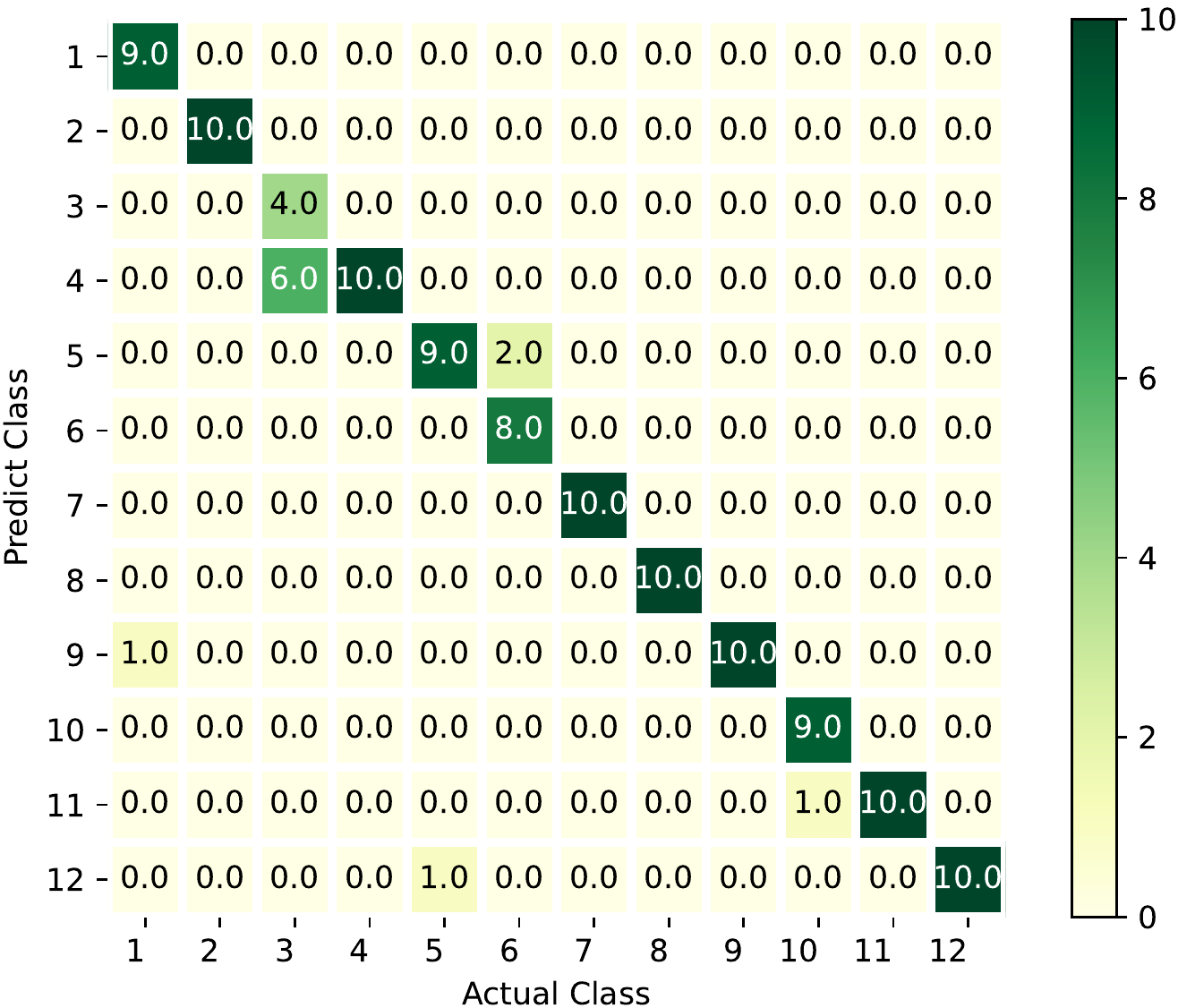}
    \caption{Confusion matrix of the vision modality.}
    \label{fig:vision matrix}
  \end{subfigure}
  \caption{Confusion matrices of single-modality and multimodality results.}
  \label{fig:matrix}
\end{figure*}


\textbf{Data pre-processing and implementation details.} Before inputting the data into the end-to-end training model shown in \cref{fig: fusion}, the data collected in \cref{sec:Testbed} needs to be preprocessed first. For each raw WiFi CSI data frame, we first remove all NaN (Not a Number) values that are caused by the loss of packet in the CSI data. Then the CSI data is normalized and sampled into a size of $3\times114\times500$ which is the input of the WiFi sensor module. 
As far as the vision data frame is concerned, the frame images extracted by the camera contain too much redundant information, which is not conducive to the extraction of gait features. As shown in \cref{fig:silhouette}, we use a Gaussian mixture model (GMM) for background subtraction to get a binarized gait silhouette~\cite{wang2003silhouette} first. Then we cut and align the silhouettes, which is the standard pipeline for vision-based gait recognition methods~\cite{takemura2018multi}. In this process, we can discard the silhouettes that do not contain any person. If the video sequence length is less than 32 frames, we repeat the last frame of the sequence to make up for 32 frames. The model structure has been illustrated in \cref{tab: GaitFi-Res}. The learning scheme of the GaitFi is implemented by PyTorch, and the model is trained on one NVIDIA GTX 1660Ti. The Adam optimizer is leveraged for better convergence. The batch size is set to 32 with a learning rate of $10^{-3}$ and a total of 30 epochs.

\begin{table}[ht]
\centering
\caption{Comparisons on real-world experiments.}
\label{tab: compare}
        \begin{tabular}{c|c|c}
            \toprule
            Method & Modality & Accuracy (\%) \\
            \midrule
            BeAware~\cite{jia2020beaware} & WiFi  & 73.3 \\
            CSAR~\cite{wang2018channel} & WiFi  & 81.7 \\
            DeepSense~\cite{zou2018deepsense}& WiFi & 85.0 \\
            CNN-LB~\cite{wu2016comprehensive} & Vision  &  68.3 \\
            PTSN~\cite{liao2017pose}  & Vision & 88.3 \\
            GaitSet~\cite{chao2019gaitset} & Vision & 92.5 \\
            \midrule
            WiFi-LRCN & WiFi & 90.8 \\ 
            LRCN & Vision & 69.2 \\
            LRCN+LSTM & Vision & 90.8 \\
            Ours (GaitFi) & WiFi+Vision & \textbf{94.2} \\
            \bottomrule
        \end{tabular}
\end{table}

\subsection{Overall Evaluation}

To evaluate the performance of the GaitFi system for the human identification task utilizing the gait recognition method, we process our dataset by utilizing methods from other research on gait recognition based on WiFi modality or vision modality. In the case of WiFi modality, we compare our method with novel WiFi-based human sensing methods including BeAware~\cite{jia2020beaware}, CSAR~\cite{wang2018channel}, DeepSense~\cite{zou2018deepsense}, and the vision-based gait recognition method including CNN-LB~\cite{wu2016comprehensive}, PTSN~\cite{liao2017pose} and GaitSet~\cite{chao2019gaitset}. In \cref{tab: compare}, our two-modality method achieves the state-of-the-art performance of 94.2\% accuracy. In comparison, the recognition accuracy of the BeAware is 73.3\%, since it only uses WiFi and a simple CNN module. When learning the WiFi modality using CSAR~\cite{wang2018channel} that consists of 4 LSTM modules, its recognition accuracy is 81.7\%. DeepSense~\cite{zou2018deepsense} innovatively combines CNN and LSTM to process WiFi CSI data, and its recognition accuracy can reach 85.0\%. For the vision modality, we first evaluate CNN-LB~\cite{wu2016comprehensive} which contains CNN feature extractors with a MLP (Multi-layer perceptron) classifier, and the recognition accuracy is 68.3\%. The PTSN~\cite{liao2017pose} proposed by Liao et al. is a very representative sequence-based gait recognition method utilizing LSTM for video gait recognition, where the recognition accuracy is 88.3\%. Then we compare our method with the state-of-the-art vision-based solution, GaitSet~\cite{chao2019gaitset}, which has outstanding recognition accuracy on the public gait dataset CASIA-B. The GaitSet achieves an accuracy of 92.5\% when it is applied to the single modality of vision in our dataset. Since the illumination condition is not ideal in the lab, the vision-based method may be affected and its performance is therefore degrading.


We also investigate our backbone network using different combinations of modalities and network structures. When we use the optimized lightweight residual convolution network WiFi-LRCN shown in \cref{tab: GaitFi-Res} for WiFi modality, the accuracy can reach 90.8\%. For vision modality, if we only use LRCN to extract the features of each frame and perform element-wise addition to get the gait features, the recognition accuracy is only 69.2\%. The reason for this is that the element-wise addition at the frame level ignores the correlation between consecutive frames in a sequence, which is important for gait. To extract sequence-level features, we use LSTM to act on the output features from LRCN, achieving 90.8\% accuracy. Although the performance of the lightweight backbone network is not as good as a complete vision-based solution GaitSet, it can save computing resources and be more efficient in identity inference. By fusing WiFi and vision two modalities, our GaitFi system achieves an accuracy rate of 94.2\%, which demonstrates the advantages of multimodal sensing. GaitFi can learn gait features of vision and WiFi modalities at the same time, improve the recognition accuracy, and enable the system to achieve better effectiveness than a single modality.

\begin{table}[t]
\centering
\caption{Experimental results under poor illumination.}
\label{tab: light}
        \begin{tabular}{c|c|c}
            \toprule
            Method & Modality & Accuracy (\%) \\
            \midrule
            BeAware~\cite{jia2020beaware} & WiFi  &  71.1\\
            CSAR~\cite{wang2018channel} & WiFi  &  74.4\\
            DeepSense~\cite{zou2018deepsense}& WiFi & 80.0 \\
            CNN-LB~\cite{wu2016comprehensive} & Vision  & 57.8 \\
            PTSN~\cite{liao2017pose}  & Vision & 62.2 \\
            GaitSet~\cite{chao2019gaitset} & Vision & 76.7 \\
            \midrule
            WiFi-LRCN & WiFi & 83.3 \\ 
            LRCN & Vision & 58.9 \\
            LRCN+LSTM & Vision &  68.9\\
            Ours (GaitFi) & WiFi+Vision & \textbf{85.6} \\
            \bottomrule
        \end{tabular}
\end{table}

\subsection{Illumination Robustness}

To study the robustness of the GaitFi system, we select 6 subjects to conduct experiments in the scene with poor illumination conditions, where 40 samples of gaits are collected from each subject, 25 of which are used as the training set and the gallery set, and the other 15 are used as the probe set. The results are shown in \cref{tab: light}. Methods based on vision modalities perform poorly. The CNN-LB achieves 57.8\%, and the PTSN only attains 62.2\%. Even the state-of-the-art vision solution, the GaitSet, only achieves 76.7\%. In contrast, the WiFi modality has better robustness against poor illumination. The BeAware, the CSAR and the DeepSense achieve 71.1\%, 74.4\%, and 80.0\%, respectively. By utilizing WiFi and vision modalities, our GaitFi system achieves the best accuracy of 85.6\%. The results illustrate that the CSI data extracted from WiFi is a good complementarity to vision modality, which can enhance the robustness of our system against poor illumination.

\subsection{Ablation Study}

\subsubsection{Modality Comparison}

We study the importance of the WiFi and vision modality when the GaitFi system conducts gait recognition. As shown in \cref{tab: Ablation}, when we only use the WiFi sensor module branch in \cref{fig: fusion}, the accuracy is only 75.0\%. Whereas, the recognition accuracy is 90.8\% when only the vision sensing module branch is used to make inferences. Both single-modality performances are lower than the 94.2\% achieved by the whole GaitFi system. These results validate that the feature fusion of WiFi and vision modality can integrate two modalities to achieve higher recognition accuracy. The confusion matrices in \cref{fig:matrix} further demonstrate the superiority of our system, and the single modality method suffers from the confusion caused by similar gender and height. It is clearly found that the wrong predictions are more likely to occur among same-gender subjects for vision modality as shown in \cref{fig:vision matrix}. For instance, samples of subject \#1 (male), subject \#3 (male), subject \#6 (female), and subject \#10 (male) are wrongly classified to subject \#9 (male), subject \#4 (male), subject \#5 (female), and subject \#11 (male), respectively. Moreover, similar heights or statures of subjects may also affect the accuracy of WiFi modality to infer human ID. In \cref{fig:wifi matrix}, when using only the WiFi modality, subjects \#1, \#2 and \#3 with similar statures are prone to confuse the model. Although the vision modality produces fewer misclassified samples than the WiFi modality, some of the misclassifications that occur in the vision modality do not occur with the WiFi modality such as subjects \#5 and \#6. Therefore, the two modalities can be the complementarity for more robust gait recognition. The multimodal result obtained in \cref{fig:fusion matrix} further demonstrates the better robustness of our system as well as the correctness of using two modalities, WiFi and vision, to sense human gait.

\begin{table}[ht]
\centering
\caption{Ablation study of different modality.}
\label{tab: Ablation}
        \begin{tabular}{c|c|c|c}
            \toprule
            WiFi module & Vision module & Accuracy (\%) & Inference time ($ms$)\\
            \midrule
            $\sqrt{}$ & $\sqrt{}$ & 94.2 & 86.3\\ \midrule
            $\sqrt{}$ &           & 75.0 & 43.1\\ \midrule
                      & $\sqrt{}$ & 90.8 & 67.6\\ \bottomrule
        \end{tabular}
\end{table}

\subsubsection{Inference Time Analysis}

To investigate the impact of multimodal learning on inference time for the GaitFi system, we calculate the inference time for one sample using the whole GaitFi system, the WiFi sensing module, and the vision sensing module, respectively. The GPU used in this experiment is only one NVIDIA GTX 1660Ti, and the results of inference time are shown in \cref{tab: Ablation}, where the inference times are $86.3$ms when using the GaitFi system, $43.1$ms when only using the WiFi sensing module and $67.6$ms when only using the vision sensing module. These results show that the multimodal learning of GaitFi only leads to marginal time consumption, which is acceptable in real-world applications.

\subsubsection{Fusion Mechanism Analysis}

\begin{table}[ht]
\centering
\caption{The effect of fusion mechanism.}
\label{tab: fusion method}
        \begin{tabular}{c|c}
            \toprule
            Fusion mechanism & Highest accuracy (\%) \\
            \midrule
            Concatenation        & 94.2 \\
            Element-wise addition    & 90.0 \\
            \bottomrule
        \end{tabular}
\end{table}

We conduct a comparative experiment on the impact of the fusion mechanism on recognition accuracy. In addition to feature concatenation in \cref{fig: fusion}, another method is to directly add the two feature vectors in each dimension numerically. In this experiment, the feature vectors extracted by the two modules are 64-dimensional, and the feature vector after the element-wise addition is still 64-dimensional, which is used to calculate the triplet loss and map to the $K$-dimensional feature space to calculate the cross-entropy loss. The vector of features obtained by element-wise addition can be represented by $z^{ij}_a$
\begin{equation}
  z^{ij}_a=z^{ij}_w+z^{ij}_v.
  \label{eq:add}
\end{equation}
\cref{tab: fusion method} shows the impact of two fusion mechanisms on the recognition accuracy, where the mean recognition accuracy of element-wise addition across three runs is only 90.0\%, not better than concatenation. These results illustrate that the feature-level concatenation has better recognition performance than the element-wise addition. The reason might be that the higher dimensional feature space has better discriminability for metric learning using the triplet loss.

\subsubsection{Loss Function Analysis}

\begin{table}[ht]
\centering
\caption{Ablation experiment of two losses.}
\label{tab: Ablation loss}
        \begin{tabular}{c|c|c}
            \toprule
            Cross-entropy loss ($\mathcal{L}_{ce}$) & Triplet loss ($\mathcal{L}_{triplet}$) & Accuracy (\%) \\
            \midrule
            $\sqrt{}$ & $\sqrt{}$ & 94.2 \\ \midrule
            $\sqrt{}$ &           & 89.2 \\ \midrule
                      & $\sqrt{}$ & 8.3 \\ \bottomrule
        \end{tabular}
\end{table}

\begin{figure}
  \centering
  \includegraphics[width=0.8\linewidth]{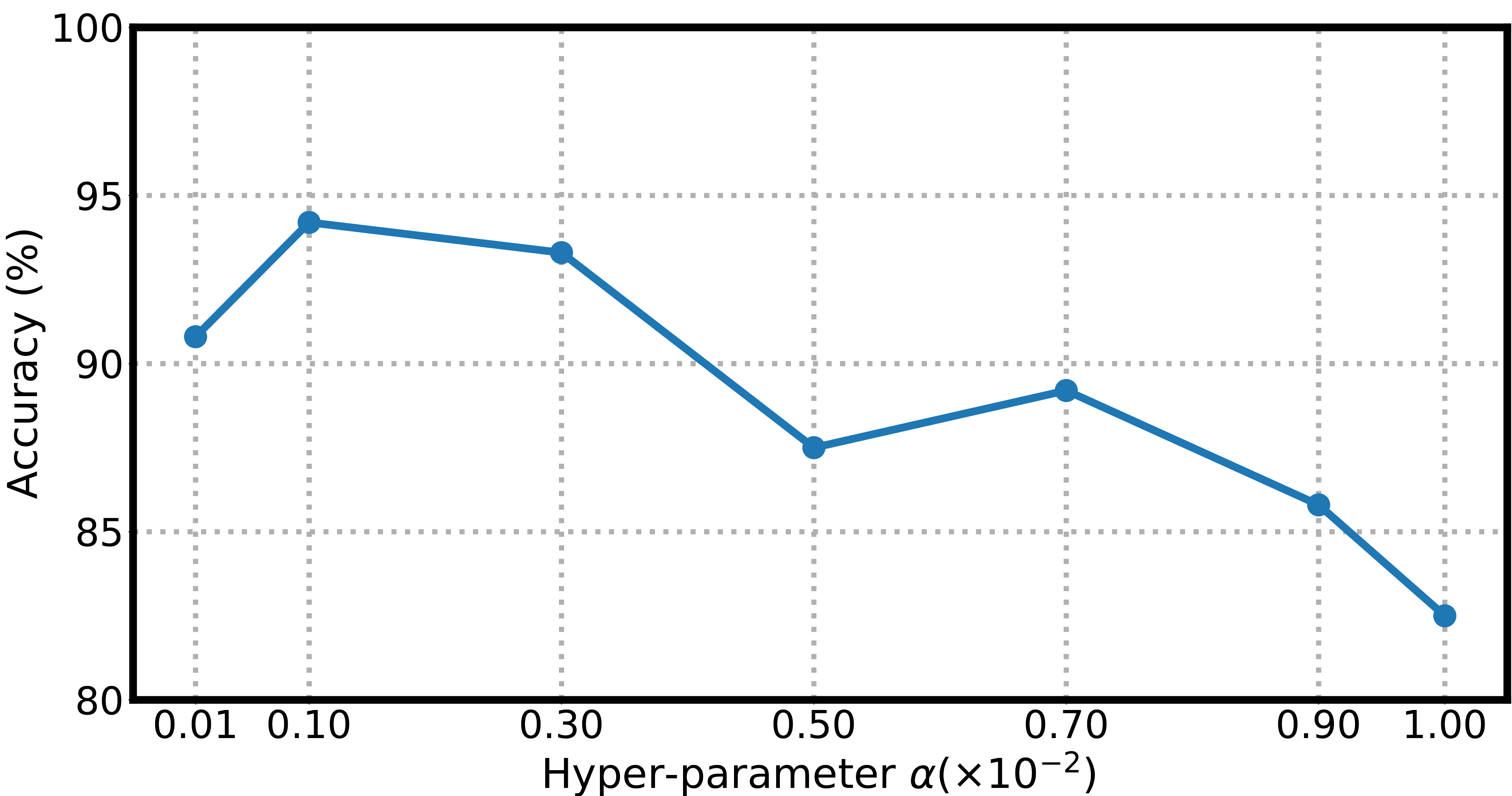}
  \caption{The impact of the hyper-parameter $\alpha$ on recognition accuracy. }
  \label{fig:alpha}
\end{figure}

The GaitFi system uses two loss functions for classification and metric learning. To test the effectiveness of two losses, we conduct ablation experiments for the loss functions, and the results are shown in \cref{tab: Ablation loss}. When only the cross-entropy loss function is used, the recognition accuracy is 89.2\%, while the accuracy of the triplet loss is only 8.3\%. It is observed that the training of the triplet loss model cannot converge. In contrast, GaitFi can achieve an accuracy of 94.2\% by using these two loss functions. The cross-entropy loss helps the model construct a discriminative feature space, and the triplet loss further refines the features to be clustered for the same subject.  This shows the two loss functions can enforce the model to learn a robust feature space, leading to better performance.

\subsubsection{Hyper-parameter Sensitivity}
In \cref{eq:loss}, we introduce the hyper-parameter $\alpha$, which is used to adjust the ratio of the cross-entropy loss and the triplet loss. To investigate the effect of $\alpha$ on the accuracy, we take different values of $\alpha$ and plot the results in \cref{fig:alpha}. It is found that a larger $\alpha$ can result in decreasing performance as the triplet loss may hinder the convergence of the cross-entropy loss, but a too low $\alpha$ makes the triplet loss not effective for feature learning. The best accuracy is achieved at $\alpha=0.001$, which is the $\alpha$ value taken in our experiments.

\section{Conclusion}
\label{sec:Conclusion}
In this paper, we propose a robust human identification system utilizing the gait recognition method that performs a multimodal fusion of WiFi signals of commercial IoT devices and videos captured by a camera through a novel deep learning method.
We firstly develop a multimodal sensing platform that can acquire WiFi CSI data from WiFi-enabled commercial off-the-shelf IoT devices and videos from cameras simultaneously. Based on residual connection, we propose LRCN, a lightweight residual convolution network to extract representative features in WiFi CSI data frames. For vision modality, a combination of LRCN and LSTM networks is used to extract representative features from visual image sequences. The extracted features of the two modalities are concatenated, and then the system performs metric learning by optimizing triplet loss and cross-entropy loss. The system makes predictions by finding the nearest neighbor of the test sample in the feature space. The experiments are conducted in the real world. According to the experimental results, the GaitFi system can achieve 94.2\% recognition accuracy, significantly outperforming other single-modal gait recognition methods based on WiFi or a camera.

\bibliographystyle{IEEEtran}
\bibliography{Reference.bib}

\begin{thebibliography}{10}
\providecommand{\url}[1]{#1}
\csname url@samestyle\endcsname
\providecommand{\newblock}{\relax}
\providecommand{\bibinfo}[2]{#2}
\providecommand{\BIBentrySTDinterwordspacing}{\spaceskip=0pt\relax}
\providecommand{\BIBentryALTinterwordstretchfactor}{4}
\providecommand{\BIBentryALTinterwordspacing}{\spaceskip=\fontdimen2\font plus
\BIBentryALTinterwordstretchfactor\fontdimen3\font minus
  \fontdimen4\font\relax}
\providecommand{\BIBforeignlanguage}[2]{{%
\expandafter\ifx\csname l@#1\endcsname\relax
\typeout{** WARNING: IEEEtran.bst: No hyphenation pattern has been}%
\typeout{** loaded for the language `#1'. Using the pattern for}%
\typeout{** the default language instead.}%
\else
\language=\csname l@#1\endcsname
\fi
#2}}
\providecommand{\BIBdecl}{\relax}
\BIBdecl

\bibitem{ali2016overview}
M.~M. Ali, V.~H. Mahale, P.~Yannawar, and A.~Gaikwad, ``Overview of fingerprint
  recognition system,'' in \emph{2016 international conference on electrical,
  electronics, and optimization techniques (ICEEOT)}.\hskip 1em plus 0.5em
  minus 0.4em\relax IEEE, 2016, pp. 1334--1338.

\bibitem{de2016iris}
M.~De~Marsico, A.~Petrosino, and S.~Ricciardi, ``Iris recognition through
  machine learning techniques: A survey,'' \emph{Pattern Recognition Letters},
  vol.~82, pp. 106--115, 2016.

\bibitem{8614364}
I.~Masi, Y.~Wu, T.~Hassner, and P.~Natarajan, ``Deep face recognition: A
  survey,'' in \emph{2018 31st SIBGRAPI Conference on Graphics, Patterns and
  Images (SIBGRAPI)}, 2018, pp. 471--478.

\bibitem{wang2021mask}
K.~Wang, S.~Wang, J.~Yang, X.~Wang, B.~Sun, H.~Li, and Y.~You, ``Mask aware
  network for masked face recognition in the wild,'' in \emph{Proceedings of
  the IEEE/CVF International Conference on Computer Vision}, 2021, pp.
  1456--1461.

\bibitem{bouchrika2018survey}
I.~Bouchrika, ``A survey of using biometrics for smart visual surveillance:
  Gait recognition,'' in \emph{Surveillance in Action}.\hskip 1em plus 0.5em
  minus 0.4em\relax Springer, 2018, pp. 3--23.

\bibitem{isaac2019trait}
E.~R. Isaac, S.~Elias, S.~Rajagopalan, and K.~Easwarakumar, ``Trait of gait: A
  survey on gait biometrics,'' \emph{arXiv preprint arXiv:1903.10744}, 2019.

\bibitem{yang2019review}
G.~Yang, W.~Tan, H.~Jin, T.~Zhao, and L.~Tu, ``Review wearable sensing system
  for gait recognition,'' \emph{Cluster Computing}, vol.~22, no.~2, pp.
  3021--3029, 2019.

\bibitem{zou2018identification}
H.~Zou, Y.~Zhou, J.~Yang, W.~Gu, L.~Xie, and C.~Spanos, ``Wifi-based human
  identification via convex tensor shapelet learning,'' pp. 1711--1719, 2018.

\bibitem{kumar2021gait}
M.~Kumar, N.~Singh, R.~Kumar, S.~Goel, and K.~Kumar, ``Gait recognition based
  on vision systems: A systematic survey,'' \emph{Journal of Visual
  Communication and Image Representation}, vol.~75, p. 103052, 2021.

\bibitem{vandersmissen2018indoor}
B.~Vandersmissen, N.~Knudde, A.~Jalalvand, I.~Couckuyt, A.~Bourdoux,
  W.~De~Neve, and T.~Dhaene, ``Indoor person identification using a low-power
  fmcw radar,'' \emph{IEEE Transactions on Geoscience and Remote Sensing},
  vol.~56, no.~7, pp. 3941--3952, 2018.

\bibitem{ding2018energy}
G.~Ding, J.~Tian, J.~Wu, Q.~Zhao, and L.~Xie, ``Energy efficient human activity
  recognition using wearable sensors,'' in \emph{2018 IEEE Wireless
  Communications and Networking Conference Workshops (WCNCW)}.\hskip 1em plus
  0.5em minus 0.4em\relax IEEE, 2018, pp. 379--383.

\bibitem{marsico2019survey}
M.~D. Marsico and A.~Mecca, ``A survey on gait recognition via wearable
  sensors,'' \emph{ACM Computing Surveys (CSUR)}, vol.~52, no.~4, pp. 1--39,
  2019.

\bibitem{chen2021attention}
S.~Chen, W.~He, J.~Ren, and X.~Jiang, ``Attention-based dual-stream vision
  transformer for radar gait recognition,'' \emph{arXiv preprint
  arXiv:2111.12290}, 2021.

\bibitem{benedek2016lidar}
C.~Benedek, B.~G{\'a}lai, B.~Nagy, and Z.~Jank{\'o}, ``Lidar-based gait
  analysis and activity recognition in a 4d surveillance system,'' \emph{IEEE
  Transactions on Circuits and Systems for Video Technology}, vol.~28, no.~1,
  pp. 101--113, 2016.

\bibitem{wang2016gait}
W.~Wang, A.~X. Liu, and M.~Shahzad, ``Gait recognition using wifi signals,'' in
  \emph{Proceedings of the 2016 ACM International Joint Conference on Pervasive
  and Ubiquitous Computing}, 2016, pp. 363--373.

\bibitem{yang2022efficientfi}
J.~Yang, X.~Chen, H.~Zou, D.~Wang, Q.~Xu, and L.~Xie, ``Efficientfi: Towards
  large-scale lightweight wifi sensing via csi compression,'' \emph{IEEE
  Internet of Things Journal}, 2022.

\bibitem{xie2015precise}
Y.~Xie, Z.~Li, and M.~Li, ``Precise power delay profiling with commodity
  wifi,'' in \emph{Proceedings of the 21st Annual International Conference on
  Mobile Computing and Networking}.\hskip 1em plus 0.5em minus 0.4em\relax ACM,
  2015, pp. 53--64.

\bibitem{yu2019review}
Y.~Yu, X.~Si, C.~Hu, and J.~Zhang, ``A review of recurrent neural networks:
  Lstm cells and network architectures,'' \emph{Neural computation}, vol.~31,
  no.~7, pp. 1235--1270, 2019.

\bibitem{yang2022benchmark}
J.~Yang, X.~Chen, D.~Wang, H.~Zou, C.~X. Lu, S.~Sun, and L.~Xie, ``Deep
  learning and its applications to wifi human sensing: A benchmark and a
  tutorial,'' \emph{arXiv preprint arXiv:2207.07859}, 2022.

\bibitem{zou2017freedetector}
H.~Zou, Y.~Zhou, J.~Yang, W.~Gu, L.~Xie, and C.~Spanos, ``Freedetector:
  Device-free occupancy detection with commodity wifi,'' in \emph{Sensing,
  Communication and Networking (SECON Workshops), 2017 IEEE International
  Conference on}.\hskip 1em plus 0.5em minus 0.4em\relax IEEE, 2017, pp. 1--5.

\bibitem{zou2018device}
H.~Zou, Y.~Zhou, J.~Yang, and C.~J. Spanos, ``Device-free occupancy detection
  and crowd counting in smart buildings with wifi-enabled iot,'' \emph{Energy
  and Buildings}, vol. 174, pp. 309--322, 2018.

\bibitem{zou2017freecount}
H.~Zou, Y.~Zhou, J.~Yang, W.~Gu, L.~Xie, and C.~Spanos, ``Freecount:
  Device-free crowd counting with commodity wifi,'' in \emph{GLOBECOM 2017-2017
  IEEE Global Communications Conference}.\hskip 1em plus 0.5em minus
  0.4em\relax IEEE, 2017, pp. 1--6.

\bibitem{zou2018deepsense}
H.~Zou, Y.~Zhou, J.~Yang, H.~Jiang, L.~Xie, and C.~J. Spanos, ``Deepsense:
  Device-free human activity recognition via autoencoder long-term recurrent
  convolutional network,'' in \emph{2018 IEEE International Conference on
  Communications (ICC)}.\hskip 1em plus 0.5em minus 0.4em\relax IEEE, 2018, pp.
  1--6.

\bibitem{zou2017multiple}
H.~Zou, Y.~Zhou, J.~Yang, W.~Gu, L.~Xie, and C.~Spanos, ``Multiple kernel
  representation learning for wifi-based human activity recognition,'' in
  \emph{2017 16th IEEE International Conference on Machine Learning and
  Applications (ICMLA)}.\hskip 1em plus 0.5em minus 0.4em\relax IEEE, 2017, pp.
  268--274.

\bibitem{zou2019wifi}
H.~Zou, J.~Yang, H.~Prasanna~Das, H.~Liu, Y.~Zhou, and C.~J. Spanos, ``Wifi and
  vision multimodal learning for accurate and robust device-free human activity
  recognition,'' in \emph{Proceedings of the IEEE/CVF Conference on Computer
  Vision and Pattern Recognition Workshops}, 2019, pp. 0--0.

\bibitem{yang2018carefi}
J.~Yang, H.~Zou, H.~Jiang, and L.~Xie, ``Carefi: Sedentary behavior monitoring
  system via commodity wifi infrastructures,'' \emph{IEEE Transactions on
  Vehicular Technology}, vol.~67, no.~8, pp. 7620--7629, 2018.

\bibitem{wang2021multimodal}
D.~Wang, J.~Yang, W.~Cui, L.~Xie, and S.~Sun, ``Multimodal csi-based human
  activity recognition using gans,'' \emph{IEEE Internet of Things Journal},
  2021.

\bibitem{wang2022caution}
------, ``Caution: A robust wifi-based human authentication system via few-shot
  open-set gait recognition,'' \emph{IEEE Internet of Things Journal}, 2022.

\bibitem{hu2022resfi}
J.~Hu, J.~Yang, J.-B. Ong, D.~Wang, and L.~Xie, ``Resfi: Wifi-enabled
  device-free respiration detection based on deep learning,'' in \emph{2022
  IEEE 17th International Conference on Control \& Automation (ICCA)}.\hskip
  1em plus 0.5em minus 0.4em\relax IEEE, 2022, pp. 510--515.

\bibitem{yang2022metafi}
J.~Yang, Y.~Zhou, H.~Huang, H.~Zou, and L.~Xie, ``Metafi: Device-free pose
  estimation via commodity wifi for metaverse avatar simulation,'' \emph{arXiv
  preprint arXiv:2208.10414}, 2022.

\bibitem{zou2018robust}
H.~Zou, J.~Yang, Y.~Zhou, L.~Xie, and C.~J. Spanos, ``Robust wifi-enabled
  device-free gesture recognition via unsupervised adversarial domain
  adaptation,'' in \emph{2018 27th International Conference on Computer
  Communication and Networks (ICCCN)}.\hskip 1em plus 0.5em minus 0.4em\relax
  IEEE, 2018, pp. 1--8.

\bibitem{yang2019learning}
J.~Yang, H.~Zou, Y.~Zhou, and L.~Xie, ``Learning gestures from wifi: A siamese
  recurrent convolutional architecture,'' \emph{IEEE Internet of Things
  Journal}, vol.~6, no.~6, pp. 10\,763--10\,772, 2019.

\bibitem{zou2018gesture}
H.~Zou, Y.~Zhou, J.~Yang, H.~Jiang, L.~Xie, and C.~J. Spanos, ``Wifi-enabled
  device-free gesture recognition for smart home automation,'' in \emph{2018
  IEEE 14th international conference on control and automation (ICCA)}.\hskip
  1em plus 0.5em minus 0.4em\relax IEEE, 2018, pp. 476--481.

\bibitem{yang2022robustsense}
J.~Yang, H.~Zou, and L.~Xie, ``Robustsense: Defending adversarial attack for
  secure device-free human activity recognition,'' \emph{arXiv preprint
  arXiv:2204.01560}, 2022.

\bibitem{yang2022autofi}
J.~Yang, X.~Chen, H.~Zou, D.~Wang, and L.~Xie, ``Autofi: Towards automatic wifi
  human sensing via geometric self-supervised learning,'' \emph{arXiv preprint
  arXiv:2205.01629}, 2022.

\bibitem{zhang2016wifi}
J.~Zhang, B.~Wei, W.~Hu, and S.~S. Kanhere, ``Wifi-id: Human identification
  using wifi signal,'' in \emph{2016 International Conference on Distributed
  Computing in Sensor Systems (DCOSS)}.\hskip 1em plus 0.5em minus 0.4em\relax
  IEEE, 2016, pp. 75--82.

\bibitem{zeng2016wiwho}
Y.~Zeng, P.~H. Pathak, and P.~Mohapatra, ``Wiwho: Wifi-based person
  identification in smart spaces,'' in \emph{2016 15th ACM/IEEE International
  Conference on Information Processing in Sensor Networks (IPSN)}.\hskip 1em
  plus 0.5em minus 0.4em\relax IEEE, 2016, pp. 1--12.

\bibitem{lv2017wii}
J.~Lv, W.~Yang, D.~Man, X.~Du, M.~Yu, and M.~Guizani, ``Wii: Device-free
  passive identity identification via wifi signals,'' in \emph{GLOBECOM
  2017-2017 IEEE Global Communications Conference}.\hskip 1em plus 0.5em minus
  0.4em\relax IEEE, 2017, pp. 1--6.

\bibitem{cao2021lightweight}
Y.~Cao, Z.~Zhou, C.~Zhu, P.~Duan, X.~Chen, and J.~Li, ``A lightweight deep
  learning algorithm for wifi-based identity recognition,'' \emph{IEEE Internet
  of Things Journal}, vol.~8, no.~24, pp. 17\,449--17\,459, 2021.

\bibitem{johansson1973visual}
G.~Johansson, ``Visual perception of biological motion and a model for its
  analysis,'' \emph{Perception \& psychophysics}, vol.~14, no.~2, pp. 201--211,
  1973.

\bibitem{wang2003silhouette}
L.~Wang, T.~Tan, H.~Ning, and W.~Hu, ``Silhouette analysis-based gait
  recognition for human identification,'' \emph{IEEE transactions on pattern
  analysis and machine intelligence}, vol.~25, no.~12, pp. 1505--1518, 2003.

\bibitem{han2005individual}
J.~Han and B.~Bhanu, ``Individual recognition using gait energy image,''
  \emph{IEEE transactions on pattern analysis and machine intelligence},
  vol.~28, no.~2, pp. 316--322, 2005.

\bibitem{xing2016complete}
X.~Xing, K.~Wang, T.~Yan, and Z.~Lv, ``Complete canonical correlation analysis
  with application to multi-view gait recognition,'' \emph{Pattern
  Recognition}, vol.~50, pp. 107--117, 2016.

\bibitem{takemura2017input}
N.~Takemura, Y.~Makihara, D.~Muramatsu, T.~Echigo, and Y.~Yagi, ``On
  input/output architectures for convolutional neural network-based cross-view
  gait recognition,'' \emph{IEEE Transactions on Circuits and Systems for Video
  Technology}, vol.~29, no.~9, pp. 2708--2719, 2017.

\bibitem{wu2016comprehensive}
Z.~Wu, Y.~Huang, L.~Wang, X.~Wang, and T.~Tan, ``A comprehensive study on
  cross-view gait based human identification with deep cnns,'' \emph{IEEE
  transactions on pattern analysis and machine intelligence}, vol.~39, no.~2,
  pp. 209--226, 2016.

\bibitem{he2018multi}
Y.~He, J.~Zhang, H.~Shan, and L.~Wang, ``Multi-task gans for view-specific
  feature learning in gait recognition,'' \emph{IEEE Transactions on
  Information Forensics and Security}, vol.~14, no.~1, pp. 102--113, 2018.

\bibitem{liao2017pose}
R.~Liao, C.~Cao, E.~B. Garcia, S.~Yu, and Y.~Huang, ``Pose-based
  temporal-spatial network (ptsn) for gait recognition with carrying and
  clothing variations,'' in \emph{Chinese conference on biometric
  recognition}.\hskip 1em plus 0.5em minus 0.4em\relax Springer, 2017, pp.
  474--483.

\bibitem{chao2019gaitset}
H.~Chao, Y.~He, J.~Zhang, and J.~Feng, ``Gaitset: Regarding gait as a set for
  cross-view gait recognition,'' in \emph{Proceedings of the AAAI conference on
  artificial intelligence}, vol.~33, no.~01, 2019, pp. 8126--8133.

\bibitem{lin2021gait}
B.~Lin, S.~Zhang, and X.~Yu, ``Gait recognition via effective global-local
  feature representation and local temporal aggregation,'' in \emph{Proceedings
  of the IEEE/CVF International Conference on Computer Vision}, 2021, pp.
  14\,648--14\,656.

\bibitem{ngiam2011multimodal}
J.~Ngiam, A.~Khosla, M.~Kim, J.~Nam, H.~Lee, and A.~Y. Ng, ``Multimodal deep
  learning,'' in \emph{ICML}, 2011.

\bibitem{li2017pixel}
S.~Li, X.~Kang, L.~Fang, J.~Hu, and H.~Yin, ``Pixel-level image fusion: A
  survey of the state of the art,'' \emph{information Fusion}, vol.~33, pp.
  100--112, 2017.

\bibitem{ross2005feature}
A.~A. Ross and R.~Govindarajan, ``Feature level fusion of hand and face
  biometrics,'' in \emph{Biometric technology for human identification II},
  vol. 5779.\hskip 1em plus 0.5em minus 0.4em\relax SPIE, 2005, pp. 196--204.

\bibitem{haghighat2016discriminant}
M.~Haghighat, M.~Abdel-Mottaleb, and W.~Alhalabi, ``Discriminant correlation
  analysis: Real-time feature level fusion for multimodal biometric
  recognition,'' \emph{IEEE Transactions on Information Forensics and
  Security}, vol.~11, no.~9, pp. 1984--1996, 2016.

\bibitem{chatzis1999multimodal}
V.~Chatzis, A.~G. Bors, and I.~Pitas, ``Multimodal decision-level fusion for
  person authentication,'' \emph{IEEE transactions on systems, man, and
  cybernetics-part a: systems and humans}, vol.~29, no.~6, pp. 674--680, 1999.

\bibitem{rahate2022multimodal}
A.~Rahate, R.~Walambe, S.~Ramanna, and K.~Kotecha, ``Multimodal co-learning:
  Challenges, applications with datasets, recent advances and future
  directions,'' \emph{Information Fusion}, vol.~81, pp. 203--239, 2022.

\bibitem{zou2019consensus}
H.~Zou, Y.~Zhou, J.~Yang, H.~Liu, H.~P. Das, and C.~J. Spanos, ``Consensus
  adversarial domain adaptation,'' in \emph{Proceedings of the AAAI Conference
  on Artificial Intelligence}, vol.~33, 2019, pp. 5997--6004.

\bibitem{ning2021review}
X.~Ning, X.~Wang, S.~Xu, W.~Cai, L.~Zhang, L.~Yu, and W.~Li, ``A review of
  research on co-training,'' \emph{Concurrency and computation: practice and
  experience}, p. e6276, 2021.

\bibitem{lin2019improving}
Y.~Lin, L.~Zheng, Z.~Zheng, Y.~Wu, Z.~Hu, C.~Yan, and Y.~Yang, ``Improving
  person re-identification by attribute and identity learning,'' \emph{Pattern
  Recognition}, vol.~95, pp. 151--161, 2019.

\bibitem{yang2018fine}
J.~Yang, H.~Zou, H.~Jiang, and L.~Xie, ``Fine-grained adaptive
  location-independent activity recognition using commodity wifi,'' in
  \emph{2018 IEEE Wireless Communications and Networking Conference
  (WCNC)}.\hskip 1em plus 0.5em minus 0.4em\relax IEEE, 2018, pp. 1--6.

\bibitem{yang2013rssi}
Z.~Yang, Z.~Zhou, and Y.~Liu, ``From rssi to csi: Indoor localization via
  channel response,'' \emph{ACM Computing Surveys (CSUR)}, vol.~46, no.~2, pp.
  1--32, 2013.

\bibitem{yang2018device}
J.~Yang, H.~Zou, H.~Jiang, and L.~Xie, ``Device-free occupant activity sensing
  using wifi-enabled iot devices for smart homes,'' \emph{IEEE Internet of
  Things Journal}, vol.~5, no.~5, pp. 3991--4002, 2018.

\bibitem{gjengset2014phaser}
J.~Gjengset, J.~Xiong, G.~McPhillips, and K.~Jamieson, ``Phaser: Enabling
  phased array signal processing on commodity wifi access points,'' in
  \emph{Proceedings of the 20th annual international conference on Mobile
  computing and networking}, 2014, pp. 153--164.

\bibitem{isufi2016autoregressive}
E.~Isufi, A.~Loukas, A.~Simonetto, and G.~Leus, ``Autoregressive moving average
  graph filtering,'' \emph{IEEE Transactions on Signal Processing}, vol.~65,
  no.~2, pp. 274--288, 2016.

\bibitem{redmon2016you}
J.~Redmon, S.~Divvala, R.~Girshick, and A.~Farhadi, ``You only look once:
  Unified, real-time object detection,'' in \emph{Proceedings of the IEEE
  conference on computer vision and pattern recognition}, 2016, pp. 779--788.

\bibitem{ahmad2021human}
T.~Ahmad, J.~Wu, I.~Khan, A.~Rahim, and A.~Khan, ``Human action recognition in
  video sequence using logistic regression by features fusion approach based on
  cnn features,'' \emph{International Journal of Advanced Computer Science and
  Applications}, vol.~12, no.~11, 2021.

\bibitem{alzubaidi2021review}
L.~Alzubaidi, J.~Zhang, A.~J. Humaidi, A.~Al-Dujaili, Y.~Duan, O.~Al-Shamma,
  J.~Santamar{\'\i}a, M.~A. Fadhel, M.~Al-Amidie, and L.~Farhan, ``Review of
  deep learning: Concepts, cnn architectures, challenges, applications, future
  directions,'' \emph{Journal of big Data}, vol.~8, no.~1, pp. 1--74, 2021.

\bibitem{He_2016_CVPR}
K.~He, X.~Zhang, S.~Ren, and J.~Sun, ``Deep residual learning for image
  recognition,'' in \emph{Proceedings of the IEEE Conference on Computer Vision
  and Pattern Recognition (CVPR)}, June 2016.

\bibitem{sheng2020deep}
B.~Sheng, F.~Xiao, L.~Sha, and L.~Sun, ``Deep spatial--temporal model based
  cross-scene action recognition using commodity wifi,'' \emph{IEEE Internet of
  Things Journal}, vol.~7, no.~4, pp. 3592--3601, 2020.

\bibitem{ioffe2015batch}
S.~Ioffe and C.~Szegedy, ``Batch normalization: Accelerating deep network
  training by reducing internal covariate shift,'' in \emph{International
  conference on machine learning}.\hskip 1em plus 0.5em minus 0.4em\relax PMLR,
  2015, pp. 448--456.

\bibitem{hermans2017defense}
A.~Hermans, L.~Beyer, and B.~Leibe, ``In defense of the triplet loss for person
  re-identification,'' \emph{arXiv preprint arXiv:1703.07737}, 2017.

\bibitem{shore1981properties}
J.~Shore and R.~Johnson, ``Properties of cross-entropy minimization,''
  \emph{IEEE Transactions on Information Theory}, vol.~27, no.~4, pp. 472--482,
  1981.

\bibitem{takemura2018multi}
N.~Takemura, Y.~Makihara, D.~Muramatsu, T.~Echigo, and Y.~Yagi, ``Multi-view
  large population gait dataset and its performance evaluation for cross-view
  gait recognition,'' \emph{IPSJ Transactions on Computer Vision and
  Applications}, vol.~10, no.~1, pp. 1--14, 2018.

\bibitem{jia2020beaware}
L.~Jia, Y.~Gu, K.~Cheng, H.~Yan, and F.~Ren, ``Beaware: Convolutional neural
  network (cnn) based user behavior understanding through wifi channel state
  information,'' \emph{Neurocomputing}, vol. 397, pp. 457--463, 2020.

\bibitem{wang2018channel}
F.~Wang, W.~Gong, J.~Liu, and K.~Wu, ``Channel selective activity recognition
  with wifi: A deep learning approach exploring wideband information,''
  \emph{IEEE Transactions on Network Science and Engineering}, vol.~7, no.~1,
  pp. 181--192, 2018.

\bibitem{yang2020mobileda}
J.~Yang, H.~Zou, S.~Cao, Z.~Chen, and L.~Xie, ``Mobileda: Toward edge-domain
  adaptation,'' \emph{IEEE Internet of Things Journal}, vol.~7, no.~8, pp.
  6909--6918, 2020.

\end{thebibliography}

\end{document}